\documentclass[10pt,twocolumn,letterpaper]{article}

\usepackage[final]{cvpr}      

\usepackage{graphicx}
\usepackage{amsmath}
\usepackage{amssymb}
\usepackage{booktabs}

\usepackage[pagebackref,breaklinks,colorlinks]{hyperref}
\usepackage[capitalize]{cleveref}

\usepackage{times}
\usepackage{epsfig}
\usepackage{multirow} 
\usepackage{color}
\usepackage{overpic}
\usepackage[T1]{fontenc}
\usepackage{hhline}
\usepackage{xcolor}
\usepackage{arydshln} 
\usepackage{pifont}
\usepackage{enumitem}
\usepackage{colortbl}
\usepackage{makecell}
\usepackage{verbatim}
\usepackage{bm}
\usepackage{caption,subcaption}

\crefname{section}{Sec.}{Secs.}
\Crefname{section}{Section}{Sections}
\Crefname{table}{Table}{Tables}
\crefname{table}{Tab.}{Tabs.}

\definecolor{darkpastelgreen}{rgb}{0.01, 0.75, 0.24}
\definecolor{darkpink}{rgb}{0.91, 0.33, 0.5}
\definecolor{mygray}{gray}{.92}
\definecolor{mygrayunder}{gray}{.98}
\definecolor{mygray1}{gray}{.50}
\definecolor{mygreen}{RGB}{93,174,86}
\definecolor{myred}{RGB}{245,55,45}
\definecolor{myorange}{RGB}{0,0,0}
\newcommand{\myPara}[1]{\textbf{#1}\quad}

\definecolor{linkcolor}{RGB}{255,0,0}
\definecolor{urlcolor}{RGB}{255,105,180}
\definecolor{citecolor}{RGB}{0, 80, 200}
\definecolor{citecolor1}{RGB}{0,153,255}
\usepackage{hyperref}
\hypersetup{colorlinks=true,linkcolor=linkcolor,urlcolor=urlcolor,citecolor=citecolor1}

\def\cvprPaperID{944} 
\def\confName{CVPR}
\def\confYear{2022}

\begin{document}

\title{Background Activation Suppression for Weakly Supervised Object Localization}

\author{Pingyu Wu\textsuperscript{1,\rm $\dagger$} \qquad
Wei Zhai\textsuperscript{1,\rm $\dagger$}\quad
Yang Cao$^{1,2,}$\thanks{{Corresponding author. $^\dagger$ Equal contributions.}}\\
$^1$ University of Science and Technology of China\\
$^2$ Institute of Artificial Intelligence, Hefei Comprehensive National Science Center\\
{\tt\small	\{wpy364755620@mail., wzhai056@mail., forrest@\}ustc.edu.cn}
}
\maketitle

\begin{abstract}
Weakly supervised object localization (WSOL) aims to localize objects using only image-level labels. Recently a new paradigm has emerged by generating a foreground prediction map (FPM) to achieve localization task. Existing FPM-based methods use cross-entropy (CE) to evaluate the foreground prediction map and to guide the learning of generator. We argue for using activation value to achieve more efficient learning. It is based on the experimental observation that, for a trained network, CE converges to zero when the foreground mask covers only part of the object region. While activation value increases until the mask expands to the object boundary, which indicates that more object areas can be learned by using activation value. In this paper, we propose a Background Activation Suppression (BAS) method. Specifically, an Activation Map Constraint module (AMC) is designed to facilitate the learning of generator by suppressing the background activation value. Meanwhile, by using the foreground region guidance and the area constraint, BAS can learn the whole region of the object. In the inference phase, we consider the prediction maps of different categories together to obtain the final localization results. Extensive experiments show that BAS achieves significant and consistent improvement over the baseline methods on the CUB-200-2011 and ILSVRC datasets. Code and models are available at \href{https://github.com/wpy1999/BAS}{\color{magenta}github.com/wpy1999/BAS}.

\end{abstract}

\begin{figure}[t]
	\centering
		\begin{overpic}[width=0.99\linewidth]{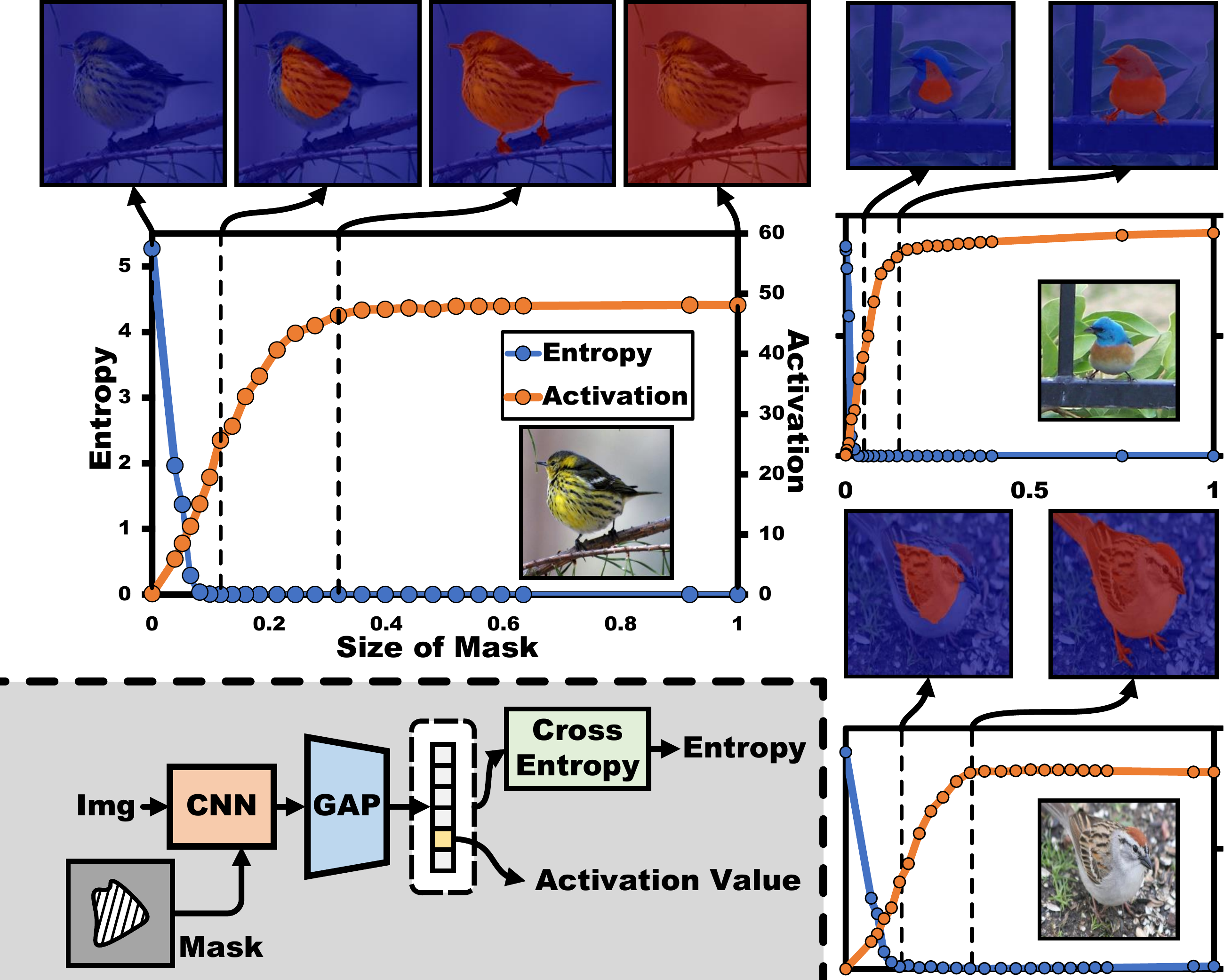}
		    \put(0., 47){\small\textbf{(A)}}
		    \put(0., 12){\small\textbf{(B)}}
	\end{overpic}
	\caption{(A) The entropy value of CE loss $w.r.t$ foreground mask and foreground activation value $w.r.t$ foreground mask. To illustrate the generality of this phenomenon, more examples are shown in the subfigure on the right. (B) Experimental procedure and related definitions. Implementation details of the experiment and further results are available in the Supplementary Material.}
	\label{fig1}
\end{figure}

\section{Introduction}

Weakly supervised object localization (WSOL) aims to identify the object's localization in a scene using only image-level labels, no bounding box annotations. WSOL is gaining more and more attention in the research community because it can visualize classification networks \cite{zhou2016learning,zhang2022vitaev2,xu2021vitae} and reduce the cost of manual labeling \cite{choe2020evaluating,yang2020combinational,gao2021ts,zhang2021weakly,kim2021normalization}. 

As an important previous work, Class Activation Map (CAM) \cite{zhou2016learning} is widely used to implement weakly supervised localization. While CAM can localize approximate object regions, it always prefers to capture the most discriminative regions rather than the overall area of the object, resulting in limited localization performance. To alleviate this problem, some methods \cite{singh2017hide,zhang2018adversarial,choe2019attention,mai2020erasing,yun2019cutmix} erase the most discriminative regions during
the training, forcing the network to learn more object features relevant to localization. \cite{zhang2018self,zhang2020inter,pan2021unveiling} are also based on CAM, which improves localization performance by establishing pixel-level spatial correlation. Additionally, some methods \cite{zhang2020rethinking,wei2021shallow,guo2021strengthen,lu2020geometry} suggest adopting a divide-and-conquer strategy to accomplish classification and localization tasks separately to avoid conflicts.

Very recently, a new paradigm \cite{meng2021foreground,xie2021online} is devised for WSOL by learning a foreground prediction map (FPM) after the feature extraction network to achieve localization without relying on CAM. Typically, ORNet \cite{xie2021online} is a two-stage approach, which first trains a classification network as an evaluator, and then utilizes CE loss to guide the learning of generator by masking the original image with foreground prediction map. In contrast to ORNet, the foreground prediction map in the FAM \cite{meng2021foreground} masks high-level information and is optimized by CE loss through two modules. In this paper, we also follow this FPM-based paradigm.

To better understand how the FPM-based paradigm works, we design the following experiments where we focus on exploring the entropy value of CE loss with respect to ($w.r.t$) foreground mask and activation value $w.r.t$ foreground mask. As shown in Fig. \ref{fig1} (A), we plot the curves of the two relationships. By observation we can find two important phenomena: 1) There is a ``mismatch'' between entropy and ground-truth mask, i.e., entropy converges to zero quickly when foreground mask retains only part of the object region. 2) There is a higher correlation between foreground activation value and foreground mask, i.e., the activation value tends to ``saturate'' when the mask expands to the object boundary. These phenomena suggest that better localization results can be achieved by using activation value compared to entropy. Moreover, from Fig. \ref{fig1} (B), it can be analyzed that CE actually facilitates the learning of the generator indirectly by influencing the activation value. Based on the above observations, a straightforward manner to obtain a complete foreground prediction map is to maximize the foreground activation value. However, considering that the maximum optimization problem is not friendly to deep neural networks, we propose to promote the learning of generation by suppressing background activation value.

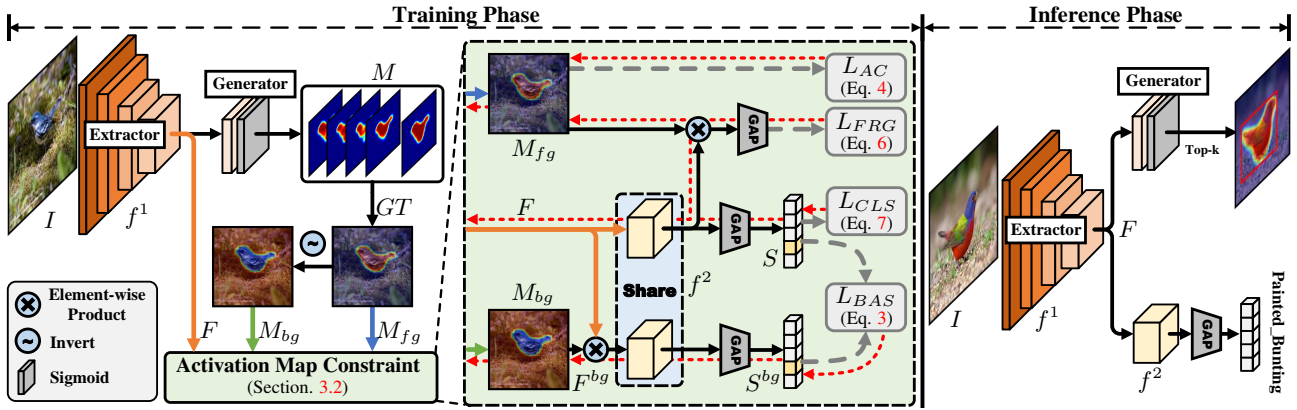
\begin{figure*}[t]
\centering
    \begin{overpic}[width=0.98\linewidth]{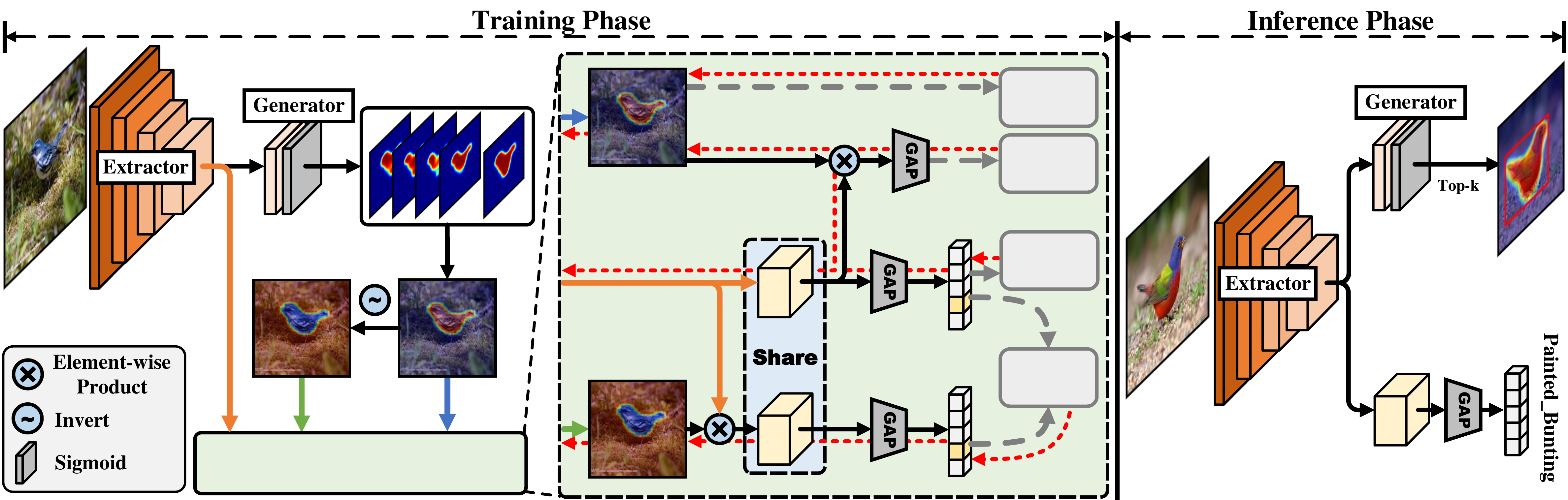}
        \put(3, 14.){\small\textbf{{$I$}}}
	    \put(65.0, 26.0){\small\textbf{${L}_{AC} $}}
	    \put(64.9, 24.4){\scriptsize{(Eq. \ref{eq:Area})}}
	    \put(64.5, 8.1){\small\textbf{${L}_{BAS}$}}
	    \put(64.9, 6.5){\scriptsize{(Eq. \ref{eq:BAS})}}
	    \put(64.7, 15.5){\small\textbf{${L}_{CLS} $} }
	    \put(64.9, 13.9){\scriptsize{(Eq. \ref{eq:CE})}}
	    \put(64.5, 21.7){\small\textbf{${L}_{FRG}$}}
	    \put(64.9, 20.1){\scriptsize{(Eq. \ref{eq:FRG})}}
	    \put(13.3, 2.8){\footnotesize\textbf{Activation Map Constraint}}
	    \put(18.8, 1.1){\scriptsize{(Section. \ref{AMC_section})}}
	    \put(28.8, 15.0){\small\textbf{{$GT$}}}
	    \put(28.8, 5.4){\small\textbf{$M_{fg}$}}
	    \put(39.3,19.6){\small\textbf{$M_{fg}$}}
	    \put(19.6, 5.4){\small\textbf{$M_{bg}$}}
	    \put(39.3,8.4){\small\textbf{$M_{bg}$}}
	    \put(28.4, 25.5){\small\textbf{{$M$}}}
	    \put(15.1, 5.4){\small\textbf{$F$}}
	    \put(39.7, 15.0){\small\textbf{$F$}}
	    \put(44.0,1.3){\small\textbf{$F^{bg}$}}
	    \put(57.5,1.3){\small\textbf{$S^{bg}$}}
	    \put(58.8,11){\small\textbf{$S$}}
	    \put(9.3,14.3){\small\textbf{$f^1$}}
	    \put(53,9){\small\textbf{$f^2$}}
	    
	    \put(73.4, 6.3){\small\textbf{{$I$}}}
	    \put(80.0, 6.3){\small\textbf{$f^1$}}
	    \put(88.0, 1.5){\small\textbf{$f^2$}}
	    \put(86.5, 13.3){\small\textbf{$F$}}
    \end{overpic}
       \caption{The architecture of the proposed BAS. In the training phase, the class-specific foreground prediction map $F^{fg}$ and the coupled background prediction map $F^{bg}$ are obtained by the generator according to the ground-truth class (GT), and then fed into the Activation Map Constraint module together with the feature map $F$. In the inference phase, we utilize Top-k to generate the final localization map.}
    \label{fig:network}
\end{figure*}

In this paper, we propose a simple but effective Background Activation Suppression (BAS) method. As shown in Figure \ref{fig:network}, BAS includes three modules: an extractor, a generator, and an Activation Map Constraint module (AMC). First, an extractor is used to extract the image features for subsequent localization and classification. The generator aims to generate a set of class-specific foreground prediction maps for localization. Then the coupled background prediction map is obtained by inversion and fed into AMC together for training. The AMC is supervised by four kinds of losses, which are background activation suppression loss, area constraint loss, foreground region guidance loss, and classification loss. The most important one is background activation suppression loss, which is devised to promote the learning of generator by minimizing the ratio of background activation and overall activation (the activation generated by the entire image). In the inference phase, we select the Top-k prediction maps to take the mean value as the final localization result based on the predicted category probabilities. Evaluations on CUB-200-2011 \cite{wah2011caltech} and ILSVRC \cite{russakovsky2015imagenet} are performed with four different types of backbones, and the experimental results show that our method achieves stable and excellent results with significant improvement over the SOTA methods. The contributions of this paper include:

\begin{itemize}
	\item [1)] This paper finds that, the essential reason why minimizing CE loss facilitates the generation of foreground maps is that it indirectly increases the foreground activation value, and accordingly proposes to facilitate the generation of foreground prediction maps by suppressing the background activation value.

	\item [2)] This paper proposes a simple but effective Background Activation Suppression (BAS) approach to facilitate the generation of foreground maps by an Activation Map Constraint (AMC) in a weakly supervised manner, which is composed of four losses including background activation suppression loss and together contribute to the generation of the foreground prediction map for localization.
    
    \item [3)] Extensive experiments on CUB-200-2011 \cite{wah2011caltech} and ILSVRC \cite{russakovsky2015imagenet} benchmarks demonstrate that our proposed method outperforms previous methods by a significant margin in terms of GT-known/Top-1/Top-5 localization.

\end{itemize}

\section{Related work}
Weakly supervised object localization (WSOL) is a challenging task that requires localizing objects using only image-level labels. To obtain the localization results from the classification network, Zhou et al.\cite{zhou2016learning} proposes to replace top layers with global average pooling, and apply the fully connected weights on depth feature maps to generate the class activation map (CAM) as the localization map. Unfortunately, CAM usually focuses on the most discriminative regions. To alleviate this problem, a type of approach proposes to use erasing strategies. HaS \cite{singh2017hide} randomly splits the original image into different patches, forcing the classification network to learn more features of objects. ACoL \cite{zhang2018adversarial} and EIL \cite{mai2020erasing} erase the region with high response in the feature map and utilize two parallel branches for adversarial erasing. Differently, ADL \cite{choe2019attention} erases the most significant regions or highlighting regions of each layer during forward propagation, to achieve a balance between classification and localization. CutMix \cite{yun2019cutmix} uses a data enhancement strategy that blends two different images for training to force the network to learn the relevant regions of different objects.

In addition, another type of approach uses the thought of spreading confidence regions to mine relevant features. SPG \cite{zhang2018self} uses thresholds to filter foreground and background regions with confidence from CAM to guide shallow network learning. Further, SPOL \cite{wei2021shallow} generates more reliable confidence regions by multiplicative feature fusion strategy, and then feeds the confidence regions into a semantic segmentation network. I2C \cite{zhang2020inter} proposes to increase the robustness and reliability of localization by considering the correlation of different pictures of the same class. Besides, SPA \cite{pan2021unveiling} uses post-processing to extract feature maps with structure-preserving. SLT \cite{guo2021strengthen} considers several similar classes as one class when 
generating classification loss and localization maps, which alleviates the problem of focusing on the most discriminative regions by increasing tolerance.

Most recently, two Foreground-Prediction-Map-based works \cite{xie2021online,meng2021foreground}, both achieve the localization task by generating a foreground prediction map. ORNet \cite{xie2021online} uses a two-stage approach, where an encode-decode layer is inserted in the shallow layer of the network as a generator and trained by the classification task in the first stage. In the second stage, the parameters of the classification network are fixed as an evaluator, and the foreground prediction map output by the generator is used to mask the image, and then fed into the evaluator for classification training, so that the foreground prediction map can learn the object region. FAM \cite{meng2021foreground} utilizes a Foreground Memory Mechanism structure to store different foreground classifiers and generate foreground prediction maps. The foreground prediction map is split into different part regions, and a class-agnostic foreground prediction map is learned by classification learning of each part region in the feature map.

It can be noticed that both ORNet \cite{xie2021online} and FAM \cite{meng2021foreground} only consider foreground regions and use cross-entropy to facilitate the learning of generator. Different from these methods, we propose a background activation suppression strategy to learn foreground prediction maps through a simple but effective approach.

\section{Methodology}
\subsection{Overview}
Based on the background activation suppression, we obtain more complete object localization maps for WSOL by proposing the BAS approach. As shown in the left subgraph of Fig. \ref{fig:network}, BAS consists of three modules: an extractor, a generator, and an Activation Map Constraint module (AMC). The extractor is used to extract features related to classification and localization. The generator is to produce the predictions of foreground maps. The AMC module is to promote the learning of extractor and generator.

We divide the original backbone network into two sub-networks $f^1$ and $f^2$ according to the location of the generator, and denote the network parameter by $\theta$. The sub-network $f^1(I,\theta ^1)$ before the generator is used as a feature extractor. Given an image $I$, the feature map $F \in R^{H \times W \times N}$ is generated by extractor in forward propagation, where $H$, $W$, and $N$ denote the height, width, and number of channels of the feature map, respectively. Afterward, the feature map $F$ is fed into the generator, which consists of a $3\times3$ convolution layer and a $Sigmoid$ activation function for generating a set of class-specific foreground prediction maps $M \in R^{H \times W \times C} $, where C is the number of dataset categories. We choose the foreground prediction map $M_{fg} \in R^{H \times W \times 1}$ corresponding to the ground-truth class and invert it to obtain the coupled background prediction map $M_{bg} \in R^{H \times W \times 1}$. Finally, $M_{fg}$, $M_{bg}$ and $F$ are fed together into AMC module for prediction map learning. We will detail describe the AMC structure and loss function in Sec. \ref{AMC_section}.

In the inference phase, as shown in the right subgraph of Fig. \ref{fig:network}. After obtained by the extractor, the feature map $F$ is fed into the generator and sub-network $f^2(F,\theta^2)$ to generate the foreground prediction maps set $M$ and the classification prediction distribution $\hat{y}$, respectively. We select the prediction maps corresponding to the Top-k categories including the ground-truth class according to the predicted category probabilities, and take their average values as the final localization results.

\subsection{Activation Map Constraint \label{AMC_section}}
The proposed AMC module utilizes foreground map, background map, and feature map as input to jointly promote the learning of extractor and generator, which is consisted of four different kinds of losses, including $\mathcal{L}_{BAS}$, $\mathcal{L}_{AC}$, $\mathcal{L}_{FRG}$, and $\mathcal{L}_{CLS}$.

\myPara{Background Activation Suppression (\bm{$\mathcal{L}_{BAS}$}).} 
For the input background prediction map $M_{bg}$, the background feature map $F^{bg}\in R^{H \times W \times C}$ is obtained by dot product with feature map $F$. Afterwards, the feature maps $F^{bg}$ and $F$ are fed to two sub-networks $f^2(F,\theta ^2)$ and $f^2(F^{bg},\theta ^2)$ with shared weights, respectively. For the sub-network with $F^{bg}$ as input, the goal is to generate the background activation value by the same function, so that this sub-network parameter is frozen in back propagation. Following the sub-network $f^2(F,\theta ^2)$ and the global average pooling (GAP) \cite{zhou2016learning}, $F$ and $F^{bg}$ produce the class probability distributions $\hat{y}$ and $\hat{y} ^ {bg}$, respectively, which can be expressed as follows:
\begin{equation}
    \label{eq:pred_y}
    \hat{y}=GAP(f^2(F,\theta ^2)),
\end{equation}
\begin{equation}
    \label{eq:pred_y_bg}
    \hat{y} ^ {bg}=GAP(f^2(F^{bg},\theta ^2)).
\end{equation}
We select the values in the $\hat{y}$ and $\hat{y} ^ {bg}$ according to the ground-truth class, denoted as activation value ${S}$ and background activation value ${S} ^ {bg}$, respectively. ${S}$ represents the activation value generated by the unmasked feature map, containing both foreground and background information, and ${S} ^ {bg}$ is the activation value generated by the background feature map, retaining only the background information. Here, we measure the difference between background activation value and activation value in a ratio form as a way to achieve background activation value suppression, and ${L}_{BAS}$ is defined in the following form:
\begin{equation}
    \label{eq:BAS}
    \mathcal{L}_{BAS}= \frac{{S} ^ {bg}}{{S} + \varepsilon},
\end{equation}
where $\varepsilon$ is a very small value ($e^{-8}$), to ensure that the equation is meaningful. 

\myPara{Area Constraint (\bm{$\mathcal{L}_{AC}$}).}
The background prediction map can be guided by $\mathcal{L}_{BAS}$ in a suppressed way, and a smaller $\mathcal{L}_{BAS}$ means that the region covered by the background prediction map is less discriminative. When the background prediction map can cover the background area well, the $\mathcal{L}_{BAS}$ it produced has to be minimal while the background area should be as large as possible, i.e., the foreground area should be as small as possible. So we use the foreground prediction map area as constraints:
\begin{equation}
    \label{eq:Area}
    \mathcal{L}_{AC}= \frac{1}{H \times W}\sum_{i}^H \sum_{j}^W M_{fg}(i,j).
\end{equation}

\myPara{Foreground Region Guidance (\bm{$\mathcal{L}_{FRG}$}).} 
Meanwhile, we retain the FPM architecture's form of using classification tasks to drive the learning of foreground prediction map, that is, using high-level semantic information to guide the foreground prediction map to the approximate correct region of the object. Therefore a foreground loss based on cross-entropy is utilized. After $F$ is fed into $f^2(F,\theta ^2)$, it is dotted with $M_{fg}$ to produce $\mathcal{L}_{FRG}$:
\begin{equation}
    \label{eq:pred_y_fg}
    \hat{y} ^ {fg}=GAP(M_{fg} \cdot f^2(F,\theta ^2)),
\end{equation}
\begin{equation}
    \label{eq:FRG}
   \mathcal{L}_{FRG}=-\sum_{i=0}^C y_{i} \log_{}{\frac{e^{\hat{y}_{i}^{fg}}}{\sum_{j}^{C} e^{\hat{y}_{j}^{fg}}}},
\end{equation}
where $y$ denotes the image-level one-hot encoding label. 

\myPara{Classification (\bm{$\mathcal{L}_{CLS}$}).}  
Besides, we obtain the classification loss $\mathcal{L}_{CLS}$ by applying cross-entropy to $\hat{y}$, which is used for classification learning of the entire image:
\begin{equation}
    \label{eq:CE}
   \mathcal{L}_{CLS}=-\sum_{i=0}^C y_{i} \log_{}{\frac{e^{\hat{y}_{i}}}{\sum_{j}^{C} e^{\hat{y}_{j}}}}.
\end{equation}
\subsection{Total loss}
By optimizing the foreground and background losses, as well as the area loss in the AMC module, can jointly guide the learning of foreground prediction map to the overall area of the object. The total loss of the BAS training process is defined in the following form:
\begin{equation}
    \label{eq:bce}
    \mathcal{L}=\mathcal{L}_{CLS} + \alpha\mathcal{L}_{FRG}  + \beta\mathcal{L}_{AC} + \lambda\mathcal{L}_{BAS},
\end{equation}
where $\alpha$, $\beta$, and $\lambda$ are hyper-parameters, $\mathcal{L}_{CLS}$ and $\mathcal{L}_{FRG}$ are both cross-entropy losses. For all backbones and datasets, we set $\lambda =1$. The ablation experiments on $\lambda$ are described in Sec. \ref{ablation_section}, and the ablation experiments on $\alpha$, $\beta$ are in Supplementary Materials.

\begin{table*}[t]
\renewcommand{\arraystretch}{1.}
\renewcommand{\tabcolsep}{8pt}
\small
\centering
\begin{tabular}{l||c||c||c|c|c||c|c|c}
\Xhline{2.0\arrayrulewidth}
\hline
\multirow{2}{*}{\textbf{Methods}} & \multirow{2}{*}{\textbf{Venue}} & \multirow{2}{*}{\textbf{Backbone}} &  \multicolumn{3}{c||}{\textbf{CUB-200-2011 \cite{wah2011caltech} Loc. Acc.}} & \multicolumn{3}{c}{\textbf{ILSVRC \cite{russakovsky2015imagenet} Loc. Acc.}} \\
\cline{4-9}
      & &  & \textbf{Top-1} & \textbf{Top-5} & \textbf{GT-known}  & \textbf{Top-1} & \textbf{Top-5} & \textbf{GT-known} \\
\hline
\Xhline{2.0\arrayrulewidth}
CAM~\cite{zhou2016learning} & CVPR$16$ &VGG16
&$41.06$&$50.66$& $55.10$  & $42.80$&$54.86$&$59.00$  \\
ACoL~\cite{zhang2018adversarial} & CVPR$18$&VGG16
&$45.92$&$56.51$& $62.96$ &$45.83$&$59.43$&$62.96$\\
ADL~\cite{choe2019attention} & CVPR$19$&VGG16
&$52.36$&- & $75.41$ &$44.92$&$-$&$-$\\
DANet~\cite{xue2019danet} & ICCV$19$ &VGG16
&$52.52$&$61.96$&$67.70$ &$-$&$-$&$-$\\
I2C~\cite{zhang2020inter} & ECCV$20$ &VGG16
&$55.99$&$68.34$&$-$ & $47.41$ & $58.51$&$63.90$\\
MEIL~\cite{mai2020erasing} & CVPR20 &VGG16
&$57.46$&$-$&$73.84$ & $46.81$ &$-$&$-$\\
GCNet~\cite{lu2020geometry} & ECCV$20$ &VGG16
&$63.24$&$75.54$&$81.10$ &$-$&$-$&$-$\\
PSOL~\cite{zhang2020rethinking} & CVPR$20$ &VGG16
& $66.30$&\underline{$84.05$}&$89.11$ & $50.89$&$60.90$&$64.03$\\
SPA~\cite{pan2021unveiling} & CVPR$21$ &VGG16
&$60.27$ &$72.50$&$77.29$ & $49.56$&$61.32$&$65.05$\\
SLT~\cite{guo2021strengthen} & CVPR$21$ &VGG16
& $67.80$&$-$&$87.60$ & $51.20$&$62.40$&$67.20$\\
FAM~\cite{meng2021foreground} & ICCV$21$ &VGG16
&\underline{$69.26$} &$-$&\underline{$89.26$} & $51.96$&$-$&\bm{$71.73$}\\
ORNet~\cite{xie2021online} & ICCV$21$ &VGG16
&$67.73$&$80.77$&$86.20$ & \underline{$52.05$}&\underline{$63.94$}&$68.27$\\
\hline
\rowcolor{mygray}
\textbf{\texttt{BAS(Ours)}}&This Work &VGG16&
\bm{$71.33 $}& \bm{$85.33 $}&\bm{$91.07 $} & \bm{$52.96 $}& \bm{$65.41 $}&\underline{$69.64 $}\\
\hline
\Xhline{1.5\arrayrulewidth}
CAM~\cite{zhou2016learning} & CVPR$16$  &MobileNetV1
&$48.07$&\underline{$59.20$}&$63.30$&$43.35$&$\underline{54.44}$&$58.97$\\
HaS~\cite{singh2017hide} & ICCV$17$ &MobileNetV1
&$46.70$&$-$&$67.31$ &$42.73$&$-$&$60.12$\\
ADL~\cite{choe2019attention} & CVPR$19$ &MobileNetV1
&$47.74$&$-$&$-$ &$43.01$&$-$&$-$\\
RCAM~\cite{bae2020rethinking} & ECCV$20$ &MobileNetV1
&$59.41$&$-$&$78.60$ &$44.78$&$-$&$61.69$\\
FAM~\cite{meng2021foreground} & ICCV$21$ &MobileNetV1
&\underline{$65.67$}&$-$&\underline{$85.71$} &\underline{$46.24$}&$-$&\underline{$62.05$}\\
\hline
\rowcolor{mygray}
\textbf{\texttt{BAS(Ours)}} & This Work & MobileNetV1
& \bm{$69.77 $}& \bm{$86.00 $}&\bm{$92.35 $} & \bm{$52.97 $}& \bm{$66.59 $}&\bm{$72.00 $}\\
\hline
\Xhline{1.5\arrayrulewidth}
CAM~\cite{zhou2016learning} & CVPR$16$ &ResNet50
&$46.71$&$54.44$&$57.35$ & $38.99$&$49.47$&$51.86$\\
ADL~\cite{choe2019attention} & CVPR$19$ &ResNet50-SE
&$62.29$&$-$&$-$ &$48.53$&$-$&$-$\\
I2C~\cite{zhang2020inter} & ECCV$20$ &ResNet50
&$-$&$-$&$-$ & $51.83$&$64.60$&$68.50$\\
PSOL~\cite{zhang2020rethinking} & CVPR$20$ &ResNet50
& $70.68$&$86.64$&$90.00$ & $53.98$&$63.08$&$65.44$\\
WTL~\cite{babar2021look} & WACV$21$ &ResNet50
&$64.70$&$-$&$77.35$ &$52.36$&$-$&$67.89$\\
FAM~\cite{meng2021foreground} & ICCV$21$ &ResNet50
&$73.74$&$-$&$85.73$ &$54.46$&$-$&$64.56$\\
SPOL~\cite{wei2021shallow} & CVPR$21$&ResNet50
& \bm{$80.12$}&\bm{$93.44$}&\bm{$96.46$} & \bm{$59.14$}&\underline{$67.15$}&\underline{$69.02$}\\
\hline
\rowcolor{mygray}
$\textbf{\texttt{BAS(Ours)}}$ & This Work &ResNet50
&\underline{$77.25 $}&\underline{$90.08 $}&\underline{$95.13 $} &\underline{$57.18 $}&\bm{$68.44 $}&\bm{$71.77 $}\\
\hline
\Xhline{1.5\arrayrulewidth}
CAM~\cite{zhou2016learning} & CVPR$16$ &InceptionV3
&$41.06$&$50.66$&$55.10$ & $46.29$&$58.19$&$62.68$\\
SPG~\cite{zhang2018self} & ECCV$18$ &InceptionV3
&$46.64$&$57.72$&$-$ &$48.60$&$60.00$&$64.69$\\
DANet~\cite{xue2019danet} & ICCV$19$ &InceptionV3
&$49.45$&$60.46$&$67.03$ &$47.53$&$58.28$&$-$\\
I2C~\cite{zhang2020inter} & ECCV$20$ &InceptionV3
&$55.99$&$68.34$&$72.60$ &$53.11$&$64.13$&$68.50$\\
GCNet~\cite{lu2020geometry} & ECCV$20$ &InceptionV3
&$58.58$&$71.00$&$75.30$ &$49.06$&$58.09$&$-$\\
PSOL~\cite{zhang2020rethinking} & CVPR$20$ &InceptionV3
& $65.51$&\underline{$83.44$}&$-$ & $54.82$&$63.25$&$65.21$\\
SPA~\cite{pan2021unveiling} & CVPR$21$ &InceptionV3
&$53.59$&$66.50$&$72.14$ &$52.73$&$64.27$&$68.33$\\
SLT~\cite{guo2021strengthen} & CVPR$21$&InceptionV3
& $66.10$&$-$&$86.50$ & \underline{$55.70$}&\underline{$65.40$}&$67.60$\\
FAM~\cite{meng2021foreground} & ICCV$21$ &InceptionV3
&\underline{$70.67$}&$-$&\underline{$87.25$} &$55.24$&$-$&\underline{$68.62$}\\
\hline
\rowcolor{mygray}
$\textbf{\texttt{BAS(Ours)}}$ & This Work &InceptionV3
&\bm{$73.29 $}&\bm{$86.31 $}&\bm{$92.24 $} &\bm{$58.51 $}&\bm{$69.00 $}&\bm{$71.93 $}\\
\hline
\Xhline{2.0\arrayrulewidth}
\end{tabular}
\caption{Comparison with state-of-the-art methods. Best results are highlighted in \textbf{bold}, second are \underline{underlined}.}
\label{table:comparsion}
\end{table*}

\begin{figure*}[t]
\centering
	\begin{overpic}[width=0.98\linewidth]{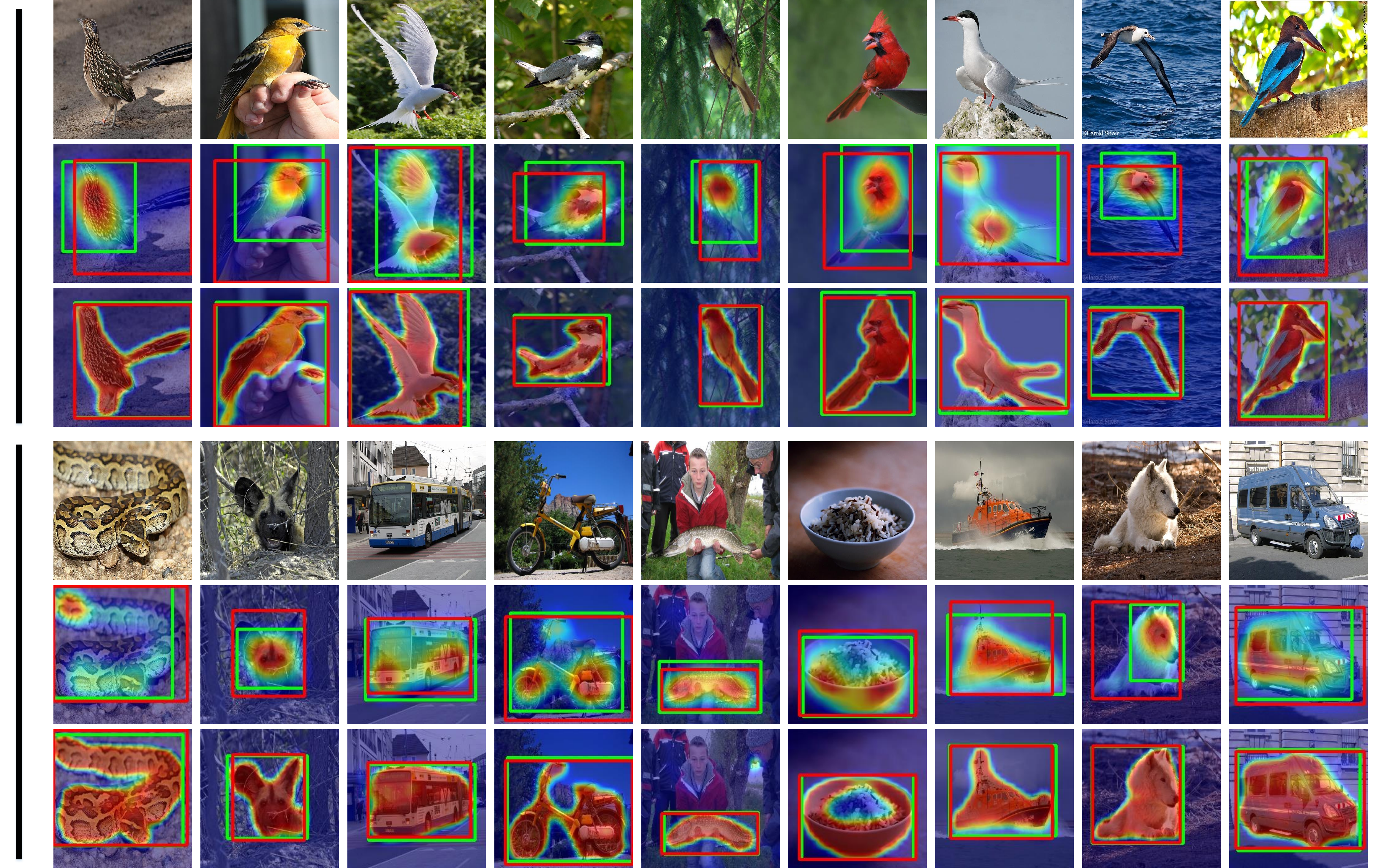}
	    \put(2., 3.5){\rotatebox{90}{\small\textbf{Ours}}}
	    \put(2., 11.5){\rotatebox{90}{\small\textbf{CAM\cite{zhou2016learning}
}}}
	    \put(2., 24.){\rotatebox{90}{\small\textbf{Image}}}
	    
	    \put(2., 35.5){\rotatebox{90}{\small\textbf{Ours}}}
	    \put(2., 43.5){\rotatebox{90}{\small\textbf{CAM\cite{zhou2016learning}
}}}
	    \put(2., 55.5){\rotatebox{90}{\small\textbf{Image}}}
	    
	    \put(-0.7, 39){\rotatebox{90}{\small\textbf{CUB-200-2011\cite{wah2011caltech}}}}
	    \put(-0.7, 11){\rotatebox{90}{\small\textbf{ILSVRC\cite{russakovsky2015imagenet}}}}
\end{overpic}
\caption{Visualization comparison with the baseline CAM \cite{zhou2016learning} method on CUB-200-2011 \cite{wah2011caltech} and ILSVRC \cite{russakovsky2015imagenet}. The ground-truth bounding boxes are in \textcolor{red}{red}, and the predictions are in \textcolor{green}{green}.}
\label{visual}
\end{figure*}

\section{Experiment}
\subsection{Experimental Setup}
\myPara{Datasets.}
We evaluate the proposed algorithm on the most popular benchmarks including CUB-200-2011 \cite{wah2011caltech} and ILSVRC \cite{russakovsky2015imagenet}. CUB-200-2011 contains 200 species of birds with $5,994$ training images and $5,794$ testing images. ILSVRC is divided into $1,000$ classes and contains about 1.2 million training images, $50,000$ validation images. Except for class labels, CUB-200-2011 also provides mask labels, which are only used to evaluate the prediction mask.

\myPara{Metrics.}
Following SPA \cite{pan2021unveiling}, we apply both bounding box and mask metrics to evaluate the performance of our BAS. For bounding box, following \cite{pan2021unveiling,zhang2020rethinking,wei2021shallow}, we use three metrics for evaluation, including GT-known localization accuracy (\textbf{GT-known Loc}), Top-1 localization accuracy (\textbf{Top-1 Loc}), and Top-5 localization accuracy (\textbf{Top-5 Loc}). Specifically, GT-known Loc is correct when the intersection over union(IoU) between the ground-truth bounding box and the predicted bounding box is greater than 0.5. Top-1/Top-5 Loc is correct when the Top-1/Top-5 predict categories contain the ground-truth class and the GT-known Loc is correct. For mask, we adopt both \textbf{Peak\underline{~}T} and \textbf{Peak\underline{~}IoU} as metrics, which are defined in SEM~\cite{zhang2020rethinking2}, to compare the prediction mask with the pixel-level ground-truth label. Peak\underline{~}IoU $\in$ $[0, 1]$ and Peak\underline{~}T $\in$ $[0, 255]$ denote the maximum intersection and its corresponding threshold, respectively. 
The larger Peak\underline{~}T indicates the higher pixel brightness value of the object area in the localization map, which can be better visualized. And a larger Peak\underline{~}IoU indicates that the localization result is closer to the target object at a specific threshold.

\myPara{Implementation Details.}
We evaluate the proposed method on the most popular backbones, including VGG16 \cite{simonyan2014very}, InceptionV3 \cite{szegedy2016rethinking}, ResNet50 \cite{he2016deep}, and MobileNetV1 \cite{howard2017mobilenets}. All networks are fine-tuned on the pre-trained weights of ILSVRC \cite{russakovsky2015imagenet}. We train 100 epochs on the CUB-200-2011 \cite{wah2011caltech} and 9 epochs on ILSVRC \cite{russakovsky2015imagenet}. In the training phase, the input images are resized to $256\times256$ and then randomly cropped to $224\times224$. When $\mathcal{L}_{BAS}$ is larger than 1, we mark it as 1, to ensure the stability of the initial training. In the inference phase, we use ten crop augmentations to get the final classification results following the settings in \cite{pan2021unveiling,guo2021strengthen,zhang2018self}. For localization, we replace the random crop with the center crop, as in previous work \cite{wei2021shallow,zhang2020rethinking,yun2019cutmix,choe2019attention}.

\subsection{Comparison with State-Of-The-Arts}
We compare the proposed BAS with state-of-the-art methods on CUB-200-2011 \cite{wah2011caltech} and ILSVRC \cite{russakovsky2015imagenet} datasets. As shown in Table \ref{table:comparsion}, BAS achieves stable and excellent performance on various backbones. On CUB-200-2011\cite{wah2011caltech}, BAS surpasses all existing methods by a large margin in terms of GT-known/Top-1/Top-5 Loc when the backbone is VGG16, MobileNetV1 and InceptionV3. Compared with the current Foreground-Prediction-Map-based method FAM \cite{meng2021foreground}, BAS achieves $6.64\%$ and $4.99\%$ GT-known Loc improvement on MobileNetV1 and InceptionV3, respectively. In addition, ResNet-BAS achieves $95.13\%$ GT-known Loc, which is a significant improvement of $9.40\%$ compared to ResNet-FAM \cite{meng2021foreground}. But compared to ResNet-SPOL \cite{wei2021shallow}, BAS is lower than it by $1.33\%$. SPOL utilizes three separated networks to achieve WSOL, first using a ResNet50 to generate class activation map, then a separate ResNet50 for segmentation, and finally an additional EfficientNet-B7\cite{tan2019efficientnet} for classification, while BAS uses only one network, which has significant advantages in efficiency. 

On ILSVRC\cite{russakovsky2015imagenet}, BAS overall exceeds all baseline methods in terms of GT-known/Top-1/Top-5 Loc on all backbones. When MobileNetV1 is used as the backbone, Our BAS achieves $72.00\%$ GT-known Loc, surpassing FAM \cite{meng2021foreground} by $9.95\%$. Moreover, InceptionV3-BAS and ResNet50-BAS obtain $71.93\%$ and $71.77\%$ GT-known Loc, respectively, establishing a novel state-of-the-art. It shows that BAS performs well on both fine-grained dataset and large universal dataset. Furthermore, we compare the localization map of the proposed BAS and CAM \cite{zhou2016learning} on CUB-200-2011 and ILSVRC in Fig. \ref{visual}. Compared to CAM, BAS can consistently cover the entire area of the object, and is sharper and more compact at the edges of the object.

\subsection{Ablation Study \label{ablation_section}}
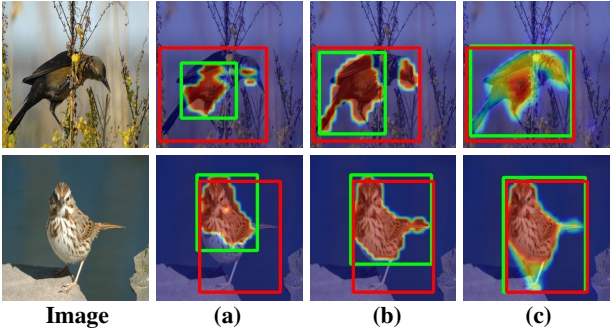
\begin{figure}[t]
\centering
    \begin{minipage}[t]{0.47\textwidth}
    \renewcommand{\arraystretch}{1.}
    \renewcommand{\tabcolsep}{3pt}
    \small
    \centering
    \begin{tabular}{c||ccc||ccc}
    \Xhline{2.\arrayrulewidth}
    \hline 
    &\textbf{Baseline} & \bm{$\mathcal{L}_{BAS}$} & \textbf{Top-k} &\textbf{Top-1}&\textbf{Top-5}&\textbf{GT-known}\\
    \Xhline{2.\arrayrulewidth}
    \hline 
    (a) &$\surd$& & &$53.45$&$65.46$&$70.39$\\
    (b) &$\surd$&$\surd$& &$69.05$&$82.43$&$87.82$ \\
    \rowcolor{mygray}
    (c) &$\surd$&$\surd$&$\surd$&\bm{$71.33$}&\bm{$85.33$}&\bm{$91.07$}  \\
    \hline
    \Xhline{2.\arrayrulewidth}
    \end{tabular}
    \end{minipage}
    \centering
    \begin{minipage}[t]{0.47\textwidth}
    \centering
    \begin{overpic}[width=0.99\linewidth]{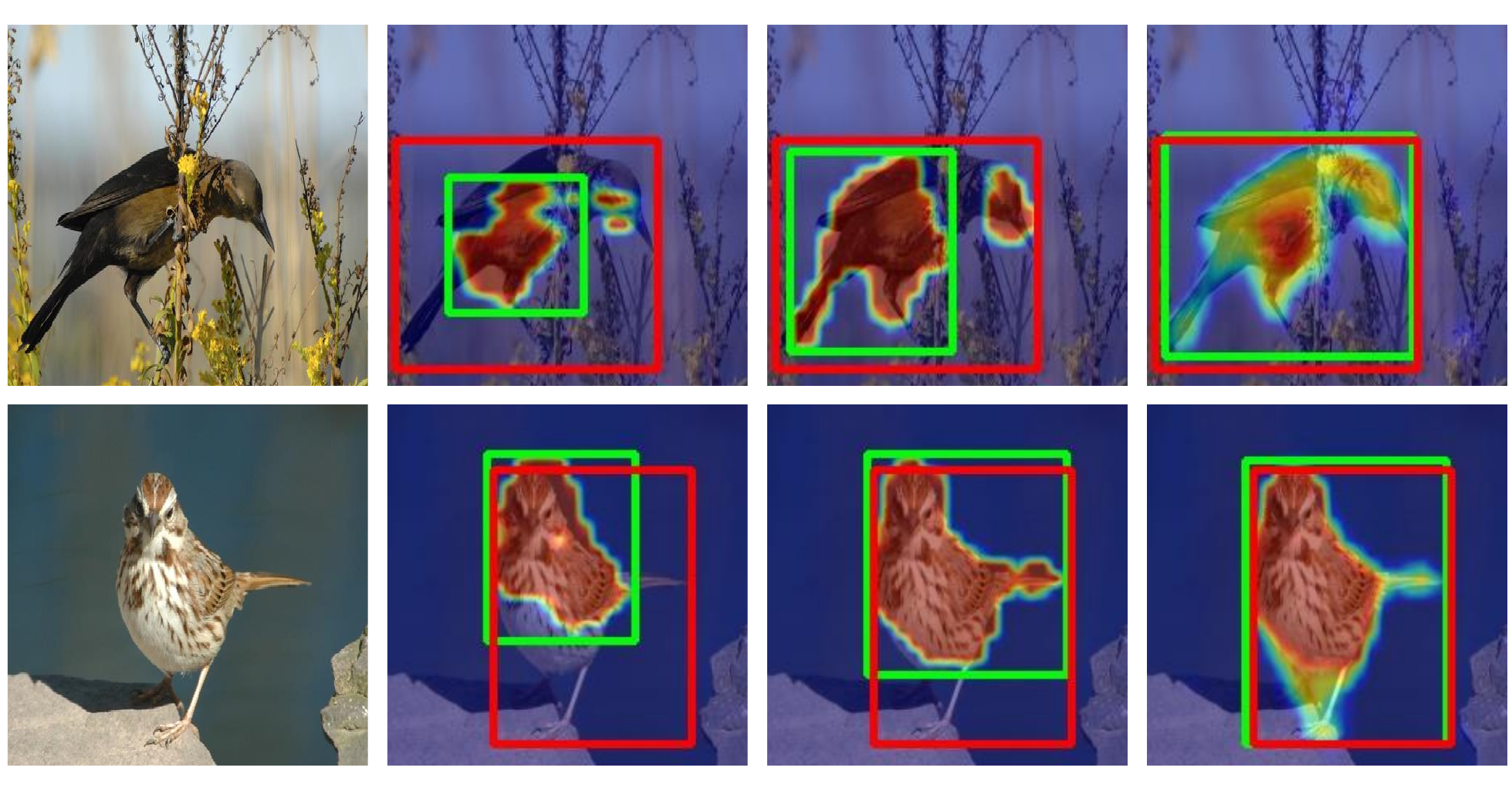}
	    \put(7.5, -1.5){\small\textbf{Image}}
	    \put(35, -1.5){\small\textbf{(a)}}
	    \put(61, -1.5){\small\textbf{(b)}}
	    \put(86, -1.5){\small\textbf{(c)}}
    \end{overpic}
    \caption{\textbf{Ablation study on our method.} (a) the baseline method. (b) add $\mathcal{L}_{BAS}$ to the baseline. (c) synthesize the final result with Top-k strategy.}
    \label{ablation}
    \end{minipage}
\end{figure}

In this section, we perform a series of ablation experiments using VGG16 \cite{simonyan2014very} as the backbone. Above all, we conduct ablation experiments on various components of BAS on CUB-200-2011 \cite{wah2011caltech}. We take $\mathcal{L}_{CLS}$, $\mathcal{L}_{FRG}$ and $\mathcal{L}_{AC}$ together as the baseline method for the Foreground-Prediction-Map-based architecture. As shown in Fig. \ref{ablation},  the addition of $\mathcal{L}_{BAS}$ based on baseline can enable the localization map to cover the object region more completely, so as to significantly improve the localization accuracy, achieving $17.43\%$ and $15.60\%$ improvement in terms of GT-known Loc and Top-1 Loc, respectively. Moreover, using Top-k strategy to integrate the final localization results, though making the localization result not as sharp as before, it can further improve the GT-known Loc (from $87.85\%$ to $91.07\%$) by increasing the connectivity of the localization map and alleviating the problem of the classification network focusing on the distinguish parts.

\begin{figure}[t]
\small
    \centering
    \begin{minipage}[t]{0.23\textwidth}
    \centering
    \begin{overpic}[width=4.05cm]{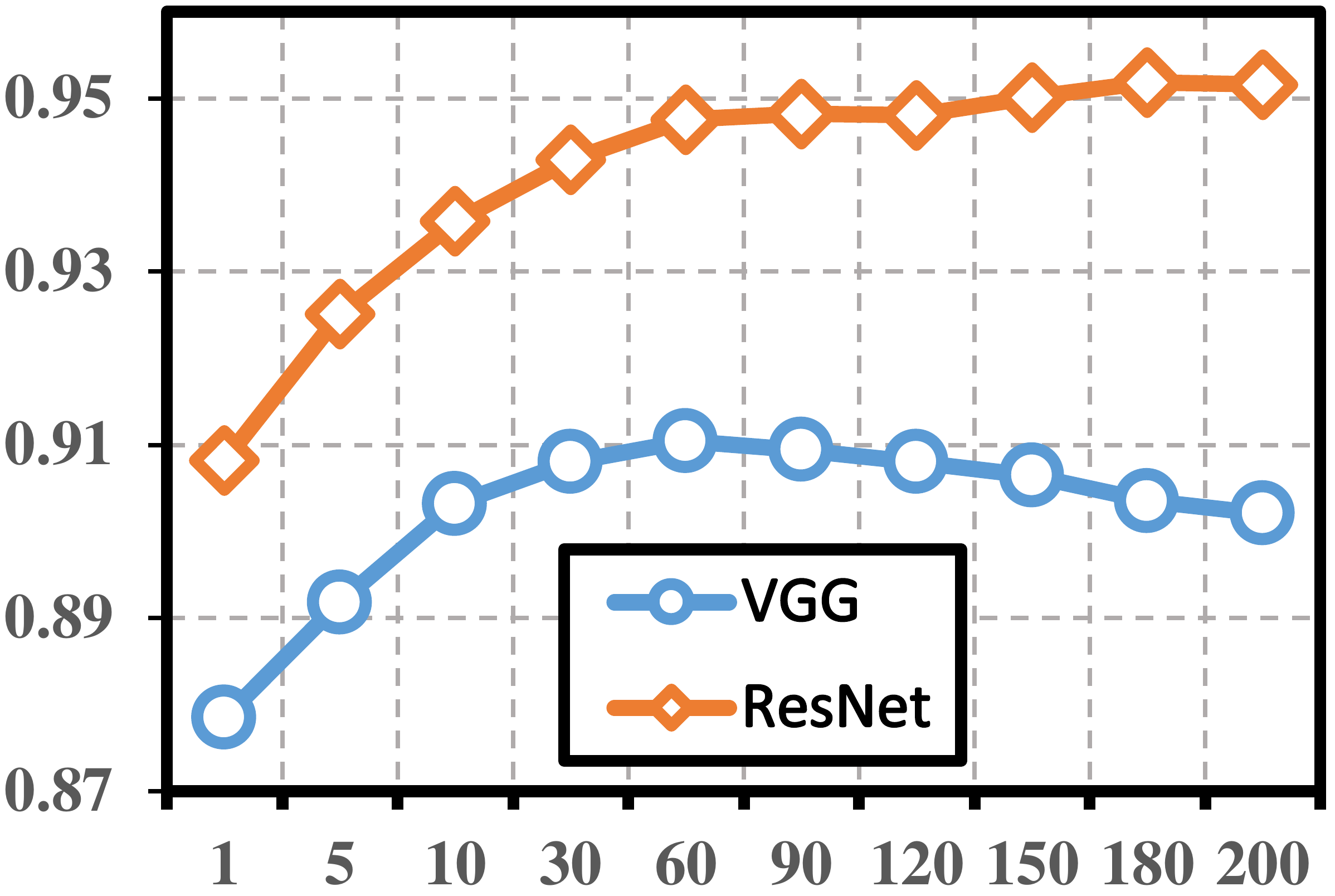}
	    \put(33,-7.){\textbf{CUB-200-2011}}
    \end{overpic}
    \end{minipage}
    \begin{minipage}[t]{0.23\textwidth}
    \centering
    \begin{overpic}[width=4.05cm]{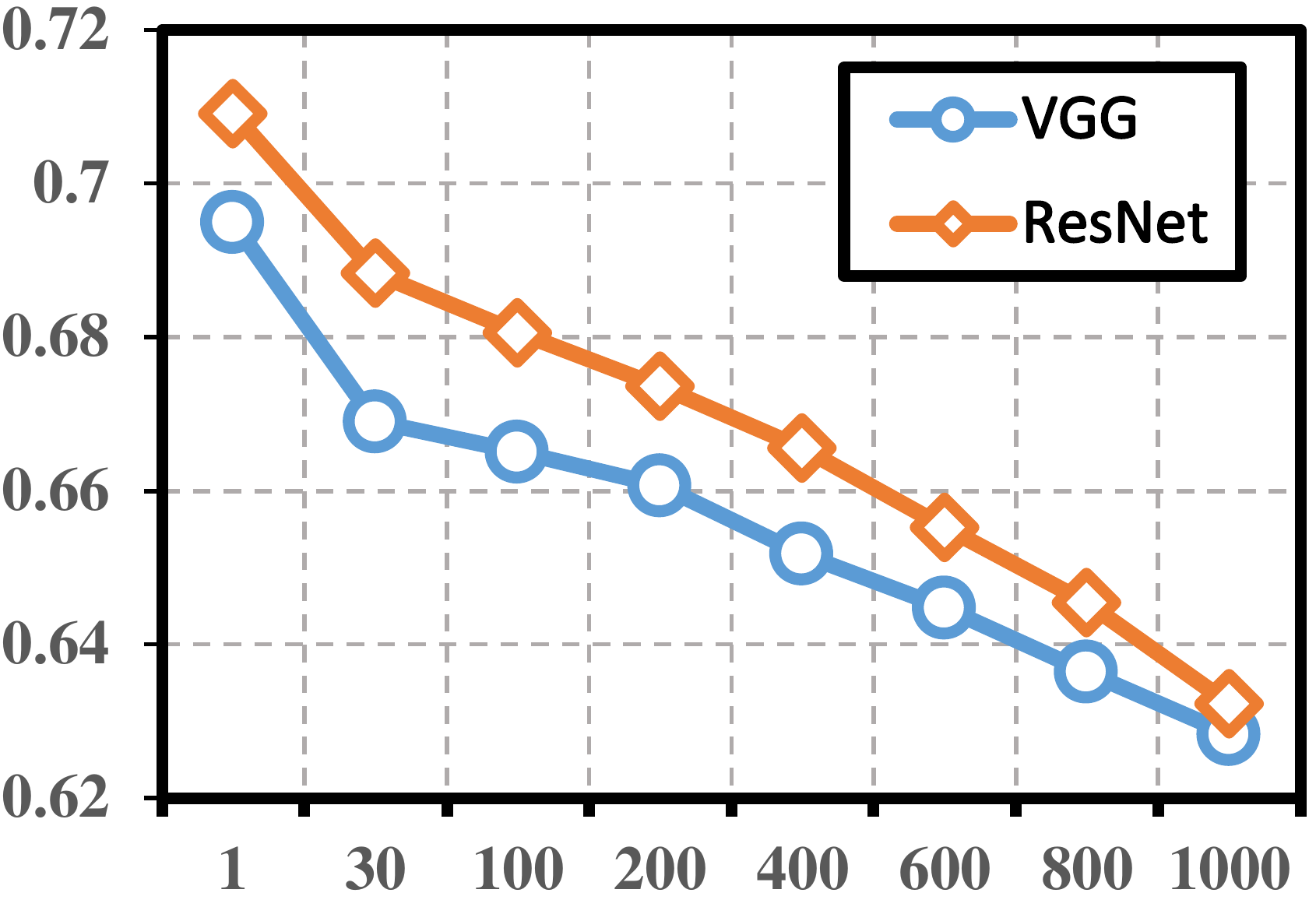}
	    \put(42,-7.){\textbf{ILSVRC}}
    \end{overpic}
    \end{minipage}
    \caption{\textbf{GT-known Loc. ($\%$) $w.r.t$ k.}}
    \label{top-k}
\end{figure}

\myPara{Hyperparameter k in Top-k strategy.}
We evaluate the effect of the hyperparameter k in our BAS. As shown in Fig. \ref{top-k}, the accuracy of GT-known Loc is improved on CUB-200-2011 when $k>1$, comparing $k=1$. For VGG16 and ResNet50, the highest localization accuracy is achieved at k of 80 and 200, respectively. It suggests that Top-k strategy can further improve the localization results by integrating the localization results of similar categories on the CUB-200-2011. In contrast, for both VGG16 and ResNet50, the best localization results are obtained for $k=1$ on ILSVRC.

\myPara{Hyperparameter $\lambda$ in total loss.}
$\lambda$ denotes the factor of $\mathcal{L}_{BAS}$. A larger $\lambda$ indicates that more regions in the prediction map are activated. As shown in Table \ref{table:lambda}, the localization accuracy continues to grow when $\lambda$ increases from 0.1 to 1, which indicates that the proposed background activation suppression strategy can significantly improve the localization accuracy. The best performance is achieved when $\lambda=1$ on CUB-200-2011.

\myPara{Generator in different layer.}
We report the result of inserting the generator at different layers of VGG16. As shown in Fig. \ref{local_strategies}, we achieve the best results by inserting the generator after the \textit{conv 4-3} layer of VGG16. When the generator learns localization information from shallow feature maps (\textit{conv3-3}), the localization map performs better at the edges of objects, but it is insufficient to resist background distractions. Generator learns localization information from the high-level feature cause imprecise localization due to the limitation of feature map resolution. 

\begin{table}[t]
\renewcommand{\arraystretch}{1}
\renewcommand{\tabcolsep}{1.5pt}
\small
\centering
\begin{tabular}{c||c|c|c|c|c|c|c|c}
\Xhline{2.\arrayrulewidth}
\hline 
\bm{$\lambda$} &\bm{$0.1$} &\bm{$0.2$} &\bm{$0.4$} &\bm{$0.6$} &\bm{$0.8$} &\bm{$1.0$} &\bm{$1.2$} &\bm{$1.4$} \\
\Xhline{2.\arrayrulewidth}
\hline 
Top-1 &$57.99$ &$68.16$ &$68.72$ &$69.99$ &$70.77$ &\bm{$71.33$} &$69.66$&$69.81$ \\
Top-5 &$69.24$ &$80.64$ &$82.77$ &$83.16$ &$84.62$ &\bm{$85.33$} &$83.95$&$83.62$ \\
GT-k. &$73.66$ &$85.87$ &$88.36$ &$89.27$ &$90.11$ &\bm{$91.07$} &$89.57$&$89.47$ \\
\hline
\Xhline{2.\arrayrulewidth}
\end{tabular}
\caption{\textbf{Performance $w.r.t$ \bm{$\lambda$}.} $\lambda$ denotes the factor of $\mathcal{L}_{BAS}$.}
\label{table:lambda}
\end{table}

\begin{figure}[t]
\centering
    \begin{minipage}[t]{0.47\textwidth}
    \renewcommand{\arraystretch}{1}
    \renewcommand{\tabcolsep}{4pt}
    \small
    \centering
    \begin{tabular}{c||c|c|c|c}
    \Xhline{2.\arrayrulewidth}
    \hline
        \textbf{Location}  & \textbf{Resolution}& \textbf{Top-1 Loc}  & \textbf{Top-5 Loc}& \textbf{GT-k. Loc} \\
    \Xhline{2.\arrayrulewidth}
    \hline
        \textit{conv 3-3}   &56$\times$56    & $65.35$  & $77.68$ & $83.38$\\
        \rowcolor{mygray}
        \textit{conv 4-3}   &28$\times$28   & $\bm{71.33}$  & $\bm{85.33}$ & $\bm{91.07}$\\
        \textit{conv 5-3}   &14$\times$14   & $66.11$  & $79.68$ & $85.36$ \\
    \hline
    \Xhline{2.\arrayrulewidth}
    \end{tabular}
    \end{minipage}
    \centering
    \begin{minipage}[t]{0.47\textwidth}
    \centering
    \begin{overpic}[width=0.99\linewidth]{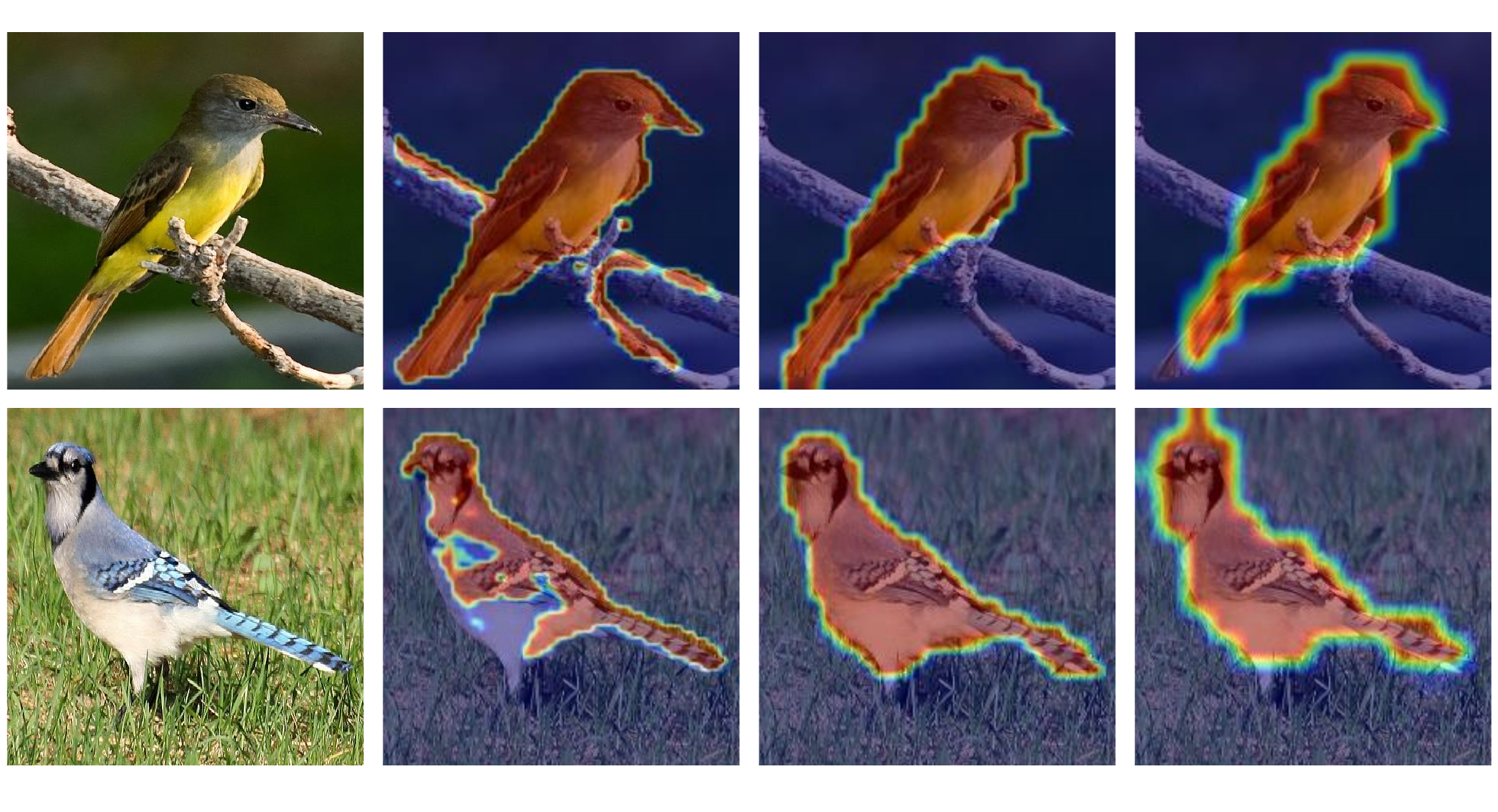}
	    \put(7.5, -1){\small\textit{Image}}
	    \put(31, -1){\small\textit{conv 3-3}}
	    \put(56, -1){\small\textit{conv 4-3}}
	    \put(80, -1){\small\textit{conv 5-3}}
    \end{overpic}
    \caption{\textbf{The results of generator in different layer.}}
    \label{local_strategies}
    \end{minipage}
\end{figure}
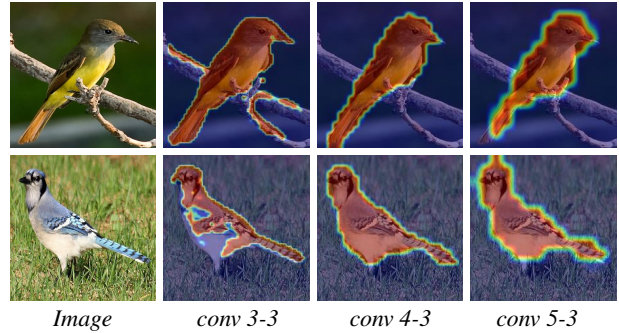

\begin{figure}[t]
\centering
    \begin{minipage}[t]{0.47\textwidth}
    \renewcommand{\arraystretch}{1}
    \renewcommand{\tabcolsep}{4pt}
    \small
    \centering
    
    \begin{tabular}{l||ccc||ccc}
    \Xhline{2.\arrayrulewidth}
    \hline 
            & \multicolumn{3}{c||}{\textbf{VGG16 Loc}} & \multicolumn{3}{c}{\textbf{ResNet50 Loc}} \\
    \Xhline{2.\arrayrulewidth}
    \hline
            \textbf{Loc.}  & \textbf{Top-1}   & \textbf{Top-5} & \textbf{GT-k.}     & \textbf{Top-1}    & \textbf{Top-5}  & \textbf{GT-k.}  \\
    \hline
    Img.    &    $70.09$     &    $83.25$   &   $88.72$     &    $73.11$  &  $86.98$  &   $92.58$     \\
    \rowcolor{mygray}
    Feat.      &    $\bm{71.33}$     &  $\bm{85.33}$   &   $\bm{91.07}$   &  $\bm{77.25}$  &    $\bm{90.08}$    &  $\bm{95.13}$ \\
   
    \hline
    \Xhline{2.\arrayrulewidth}
    \end{tabular}
    \end{minipage}
    \centering
    \begin{minipage}[t]{0.47\textwidth}
    \centering
    \begin{overpic}[width=0.99\linewidth]{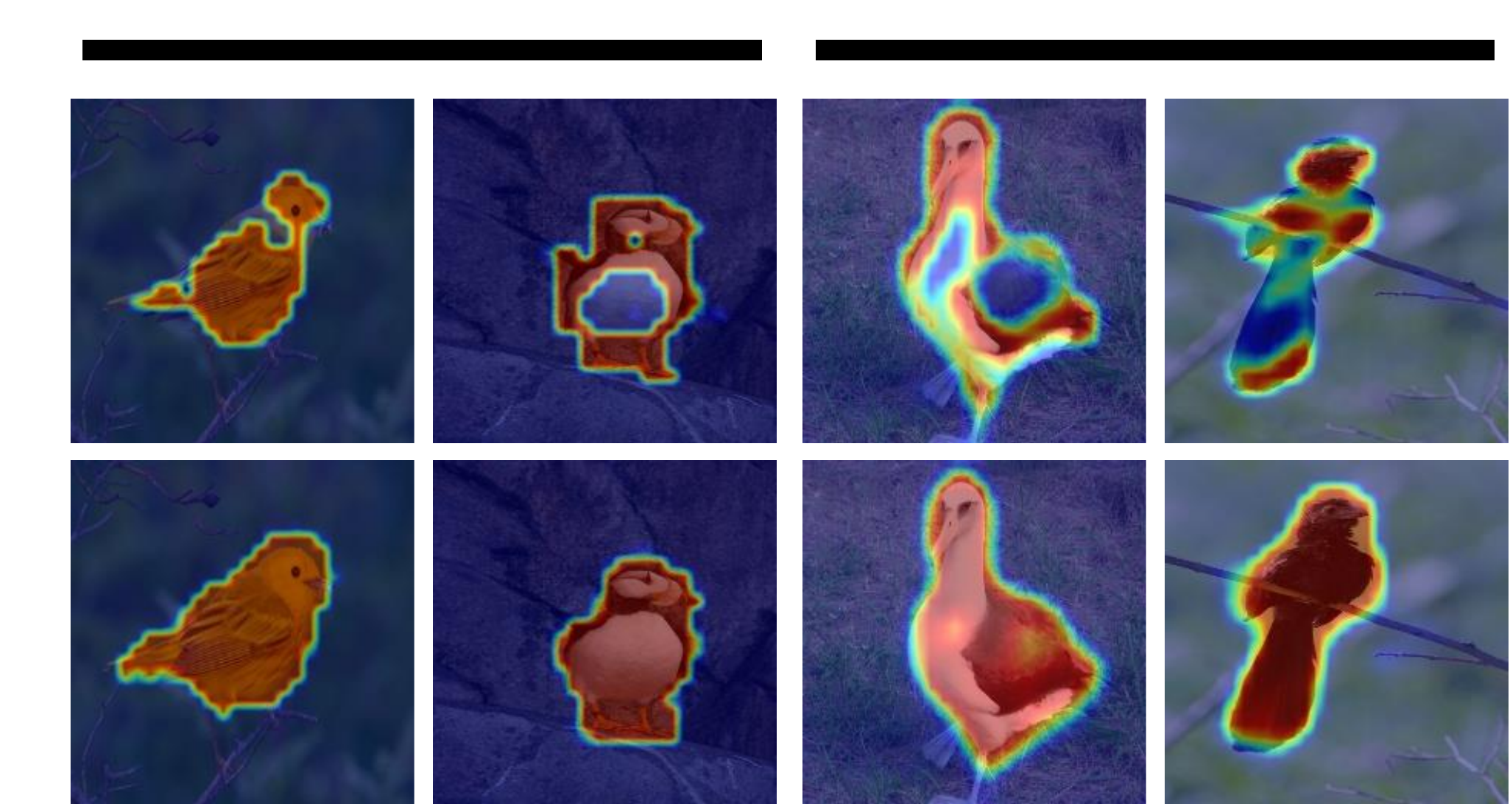}
	    \put(-0.4, 29.5){\rotatebox{90}{\small\textbf{Image}}}
	    \put(-0.4, 5){\rotatebox{90}{\small\textbf{Feature}}}
	    \put(21.,49.2){\colorbox{white}{{\color{black} \small\textbf{VGG16}}}}
	    \put(67.,49.2){\colorbox{white}{{\color{black} \small\textbf{ResNet50}}}}
    \end{overpic}
    \caption{\textbf{Comparison of background prediction maps learned from the original image $vs$ the feature map.}}
    \label{erase_strategies}
    \end{minipage}
\end{figure}
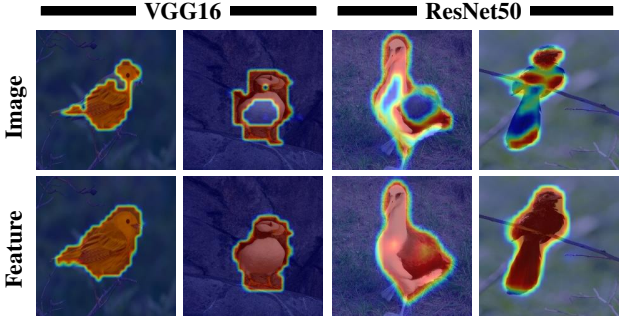

\myPara{Original image $vs$ feature map.}
We conduct experiments on the intervention position of the background prediction map (original image $vs$ feature map). As shown in Fig. \ref{erase_strategies}, we note that the masked feature map approach achieves higher accuracy and better coverage of the localization results on the object, while the results generated in the original map focus more on the edge texture of the object. It may be due to the fact that the learning process in shallow layers usually focuses on common basic features (e.g., edges, textures) and ignores high-level features.

\subsection{Performance Analysis}
\begin{figure}[t]
\small
    \centering
    \begin{minipage}[t]{0.23\textwidth}
    \centering
    \begin{overpic}[width=4.05cm]{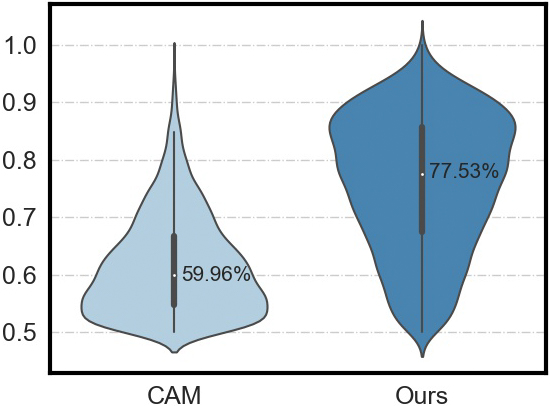}
	    \put(30,-7.){\textbf{CUB-200-2011}}
    \end{overpic}
    \end{minipage}
    \begin{minipage}[t]{0.23\textwidth}
    \centering
    \begin{overpic}[width=4.05cm]{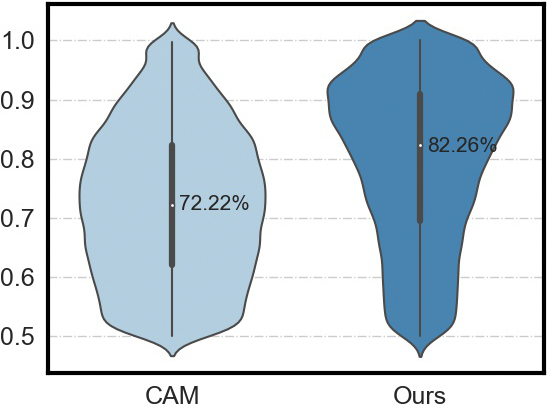}
	    \put(39,-7.){\textbf{ILSVRC}}
    \end{overpic}
    \end{minipage}
    \caption{\textbf{Statistical analysis of correct bounding boxes.}}
    \label{localization_iou}
\end{figure}

\myPara{Localization Quality.}
In Fig.\ref{localization_iou}, we show the statistical analysis of the IoU between the bounding boxes and the ground-truth boxes when localized correctly, following DANet \cite{xue2019danet}. On CUB-200-2011, we achieve $77.53\%$ IoU median when localized correctly, exceeding CAM \cite{zhou2016learning} by $17.57\%$, and correspondingly by $10.04\%$ on ILSVRC. From the median IoU and the IoU distribution, it can be seen that the proposed BAS significantly improves the localization quality on both datasets.

\begin{figure}[t]
\footnotesize
\centering
\begin{minipage}{0.62\linewidth}
    \centering
    \begin{overpic}[width=4.45cm]{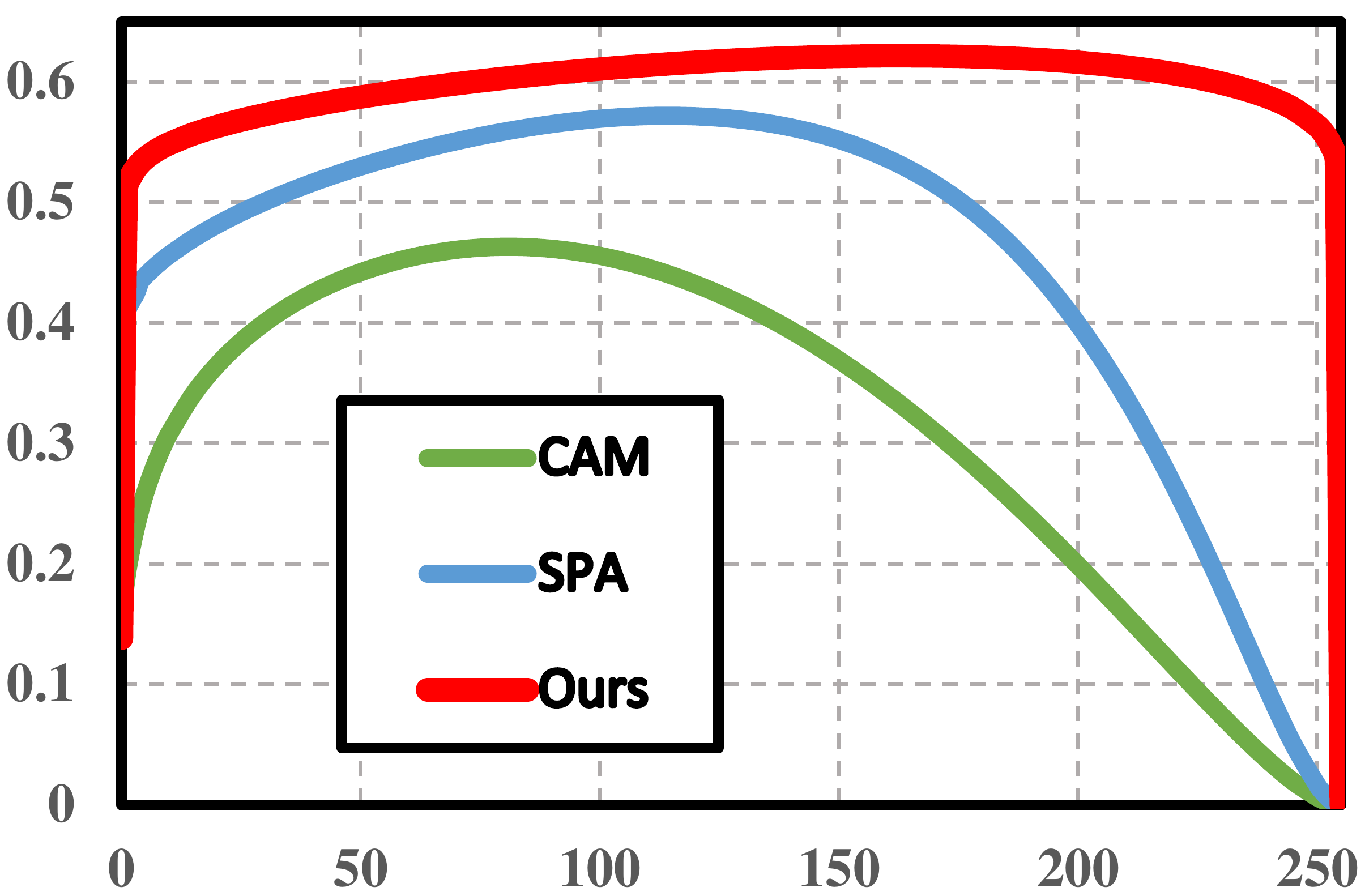}
	    \put(40,-6.){\textbf{Threshold}}
	    \put(-5,30.){\rotatebox{90}{\textbf{IoU}}}
    \end{overpic}
\end{minipage}
\hfill
\centering
\begin{minipage}{.37\linewidth}
    \renewcommand{\arraystretch}{1}
    \renewcommand{\tabcolsep}{1.pt}
    \small
    \centering
    \begin{tabular}{l||c|c}
    \Xhline{2.\arrayrulewidth}
    \hline
    \textbf{Methods}& \textbf{P.-T} & \textbf{P.-IoU} \\
    \Xhline{2.\arrayrulewidth}
    \hline
    CAM\cite{zhou2016learning}    & $80$     & $46.31$    \\
    SPA\cite{pan2021unveiling}   & $120$    & $57.13$    \\
    \rowcolor{mygray}
    BAS & $\bm{150}$    & $\bm{62.10}$   \\
    \hline
    \Xhline{2.\arrayrulewidth}
    \end{tabular}
\end{minipage}
    \caption{\textbf{Segmentation Quality.} IoU-Threshold curves for different baseline methods and evaluation results of Peak-T, Peak-IoU on CUB-200-2011.}
    \label{segmentation_iou}
\end{figure}

\myPara{Segmentation Quality.}
We compare the localization map with the ground-truth mask label using two metrics, Peak\underline{~}T and Peak\underline{~}IoU, following SPA \cite{pan2021unveiling}. As shown in Fig. \ref{segmentation_iou}, we evaluate the performance of the proposed BAS with CAM \cite{zhou2016learning} and SPA\cite{pan2021unveiling} on VGG16. Compared to SPA, BAS achieves significant and consistent improvement on both Peak\underline{~}T and Peak\underline{~}IoU, with a $4.97\%$ improvement in Peak\underline{~}IoU and 30 in Peak\underline{~}T, respectively. And it can be seen from the left subgraph of Fig. \ref{segmentation_iou}, our IoU-Threshold curve covers a larger area, which indicates that the localization map produced by BAS has fewer low confidence regions and is closer to the original object region.

\subsection{Limitation}
\begin{figure}[t]
    \centering
    \begin{minipage}[t]{0.23\textwidth}
    \centering
    \begin{overpic}[width=4.05cm]{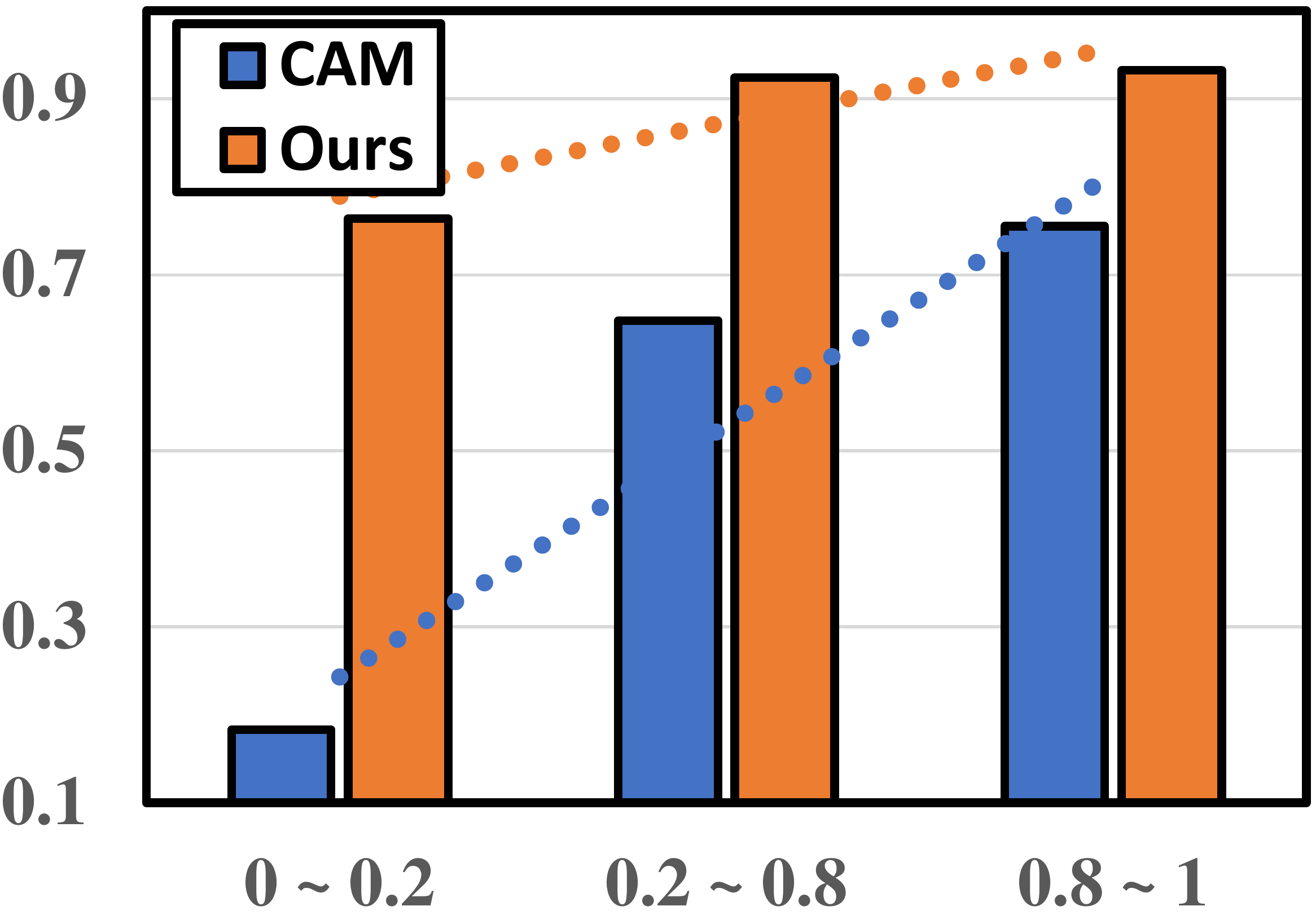}
	    \put(31.0,-7.){\small{CUB-200-2011}}
    \end{overpic}
    \end{minipage}
    \begin{minipage}[t]{0.23\textwidth}
    \centering
    \begin{overpic}[width=4.05cm]{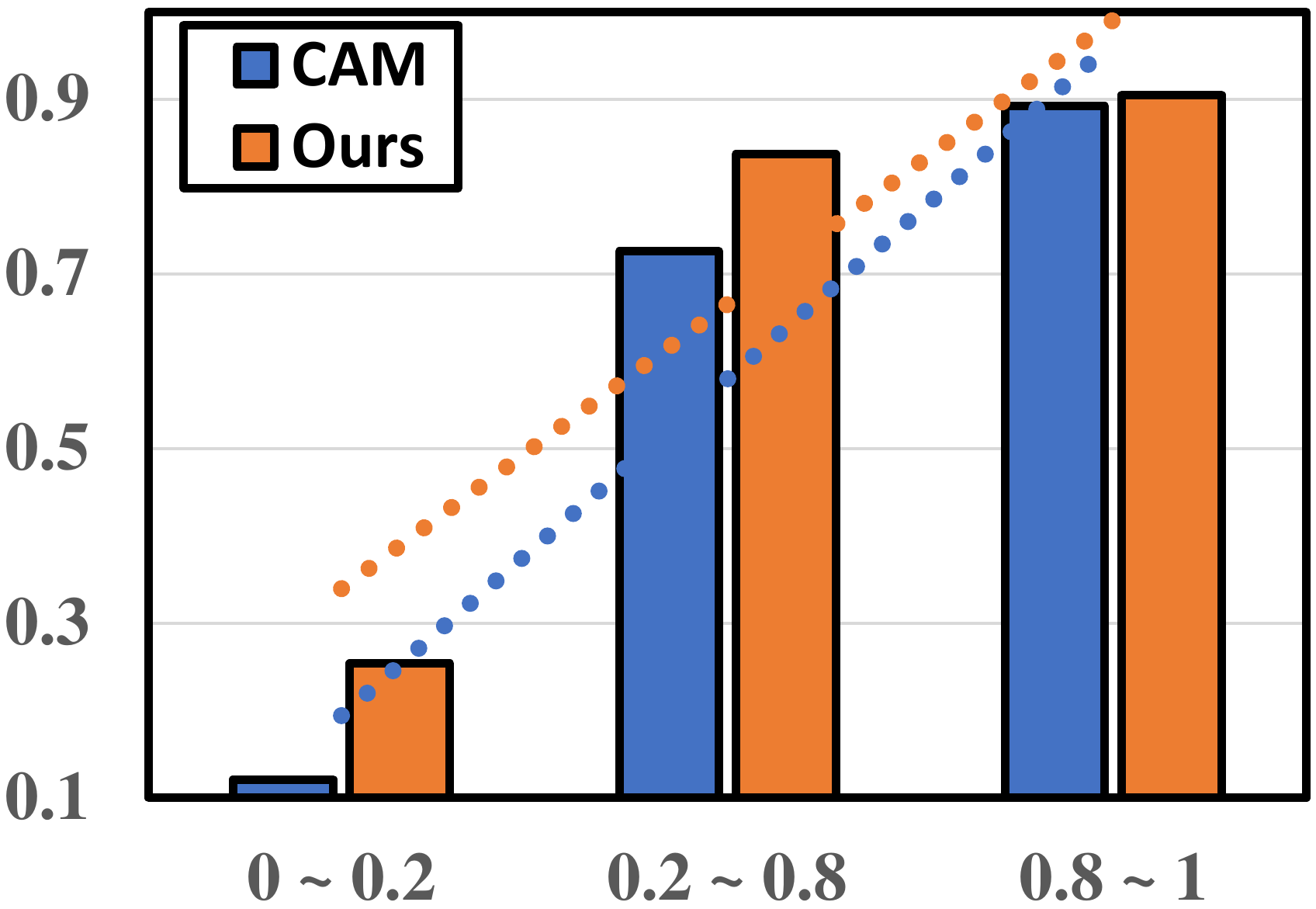}
	    \put(41.0,-7.){\small{ILSVRC}}
    \end{overpic}
    \end{minipage}
    \caption{\textbf{Limitation.} The performance $w.r.t$ different scale. The test set is divided into three intervals of 0\textasciitilde0.2, 0.2\textasciitilde0.8, 0.8\textasciitilde1 according to the percentage of ground-truth box area compared to the image area, and statistics on the localization accuracy of BAS and CAM for various sizes of objects.}
    \label{limitation}
\end{figure}

In this section, we discuss the limitation of BAS. We split the dataset according to the ground-truth box size, as shown in Fig. \ref{limitation}. We note that BAS is inconsistent for localizing objects of different sizes, with poorer localization ability for small objects, especially on ILSVRC. Although it is a great improvement over CAM \cite{zhou2016learning}, it is still a challenge to locate small objects better, mainly due to unbalanced distribution between features of the foreground and background. To overcome this limitation, we believe that in further work, object location can be achieved in two stages. Based on the fact that WSOL works better for localizing large objects, we can determine the approximate region of the objects in the first stage, and then crop and resize the corresponding region to convert the original small objects into a larger one, thereby performing localization in the second stage. 

\section{Conclusion}

In this paper, we find previous FPM-based work using cross-entropy to facilitate the learning of foreground prediction maps, essentially by changing the activation value, and the activation value shows a higher correlation with the foreground mask. Thus, we propose a Background Activation Suppression (BAS) approach to promote the generation of foreground maps by an Activation Map Constraint (AMC) module, which facilitates the learning of foreground prediction maps mainly through the suppression of background activation. Extensive experiments on CUB-200-2011 and ILSVRC verify the effectiveness of the proposed BAS, which surpasses previous methods by a large margin.

\myPara{Societal Implications.}
This work may have the following societal Implications. Achieving object localization without the need for location annotations, which will largely reduce manual labeling costs. This is especially valuable for industry or in the medical field since labeling is costly.

\textbf{Acknowledgments.} Supported by National Key R\&D Program of China under Grant 2020AAA0105701, National Natural Science Foundation of China (NSFC) under Grants 61872327 and Major Special Science and Technology Project of Anhui (No. 012223665049).


{\small
\bibliographystyle{IEEEtran}
\bibliography{main.bib}
}

\clearpage

\appendix
\begin{figure*}[h!]
	\centering
		\begin{overpic}[width=0.95\linewidth]{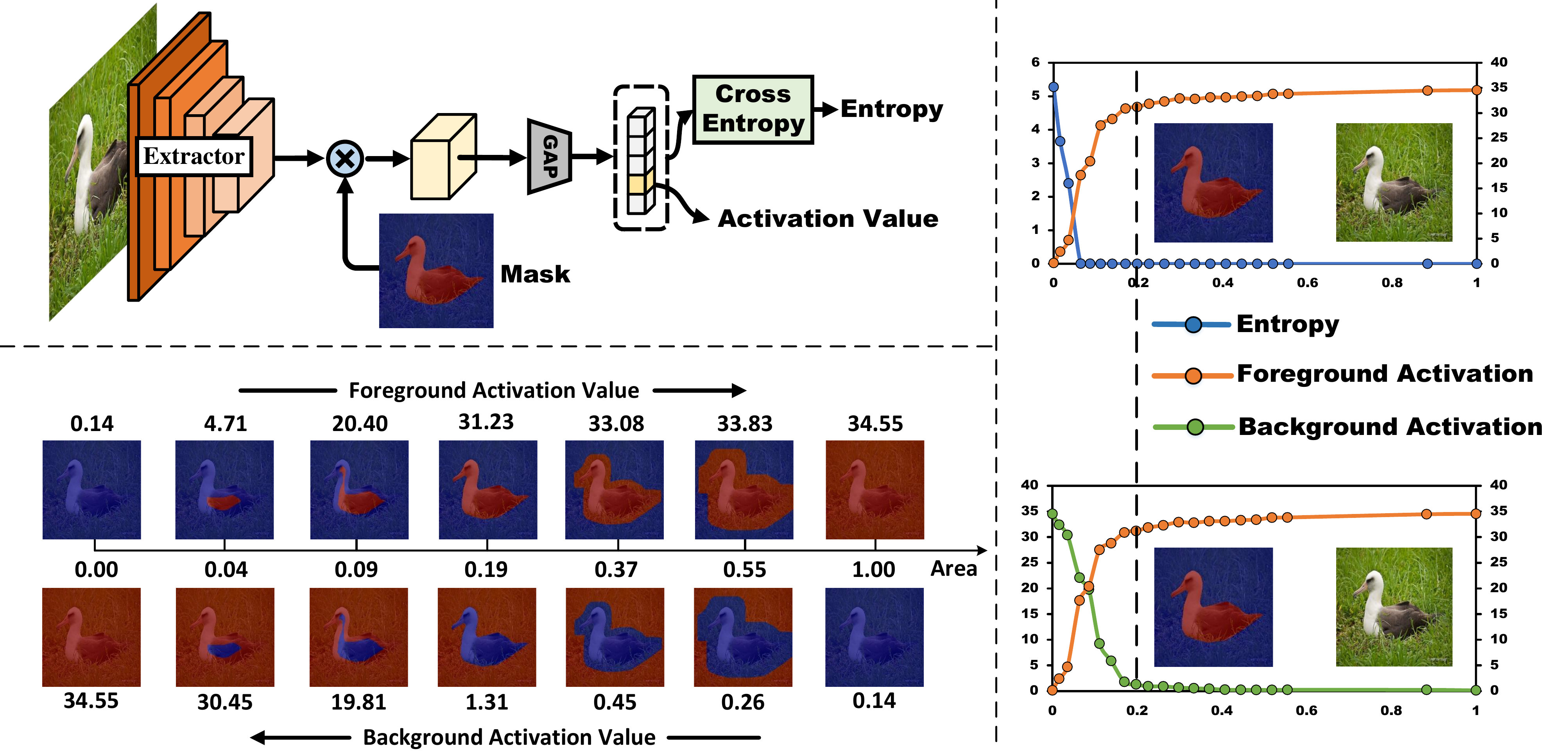}
		    \put(0, 28){\small\textbf{(A)}}
		    \put(0, 0){\small\textbf{(B)}}
		    \put(66., 23){\small\textbf{(C)}}
	\end{overpic}
	\caption{(A) Flow diagram of experiment. The activation value and cross-entropy corresponding to the mask are generated by masking the feature map. (B) The foreground activation value and background activation value are obtained by reserving the foreground area and background area, respectively. (C) The curve of entropy, foreground activation, and background activation with mask area, the dashed line represents the position of the ground-truth mask.}
	\label{mask_network}
\end{figure*}
\section*{Supplementary Materials}

\section{Exploratory Experiment}
\begin{figure*}[h]
	\centering
		\begin{overpic}[width=0.95\linewidth]{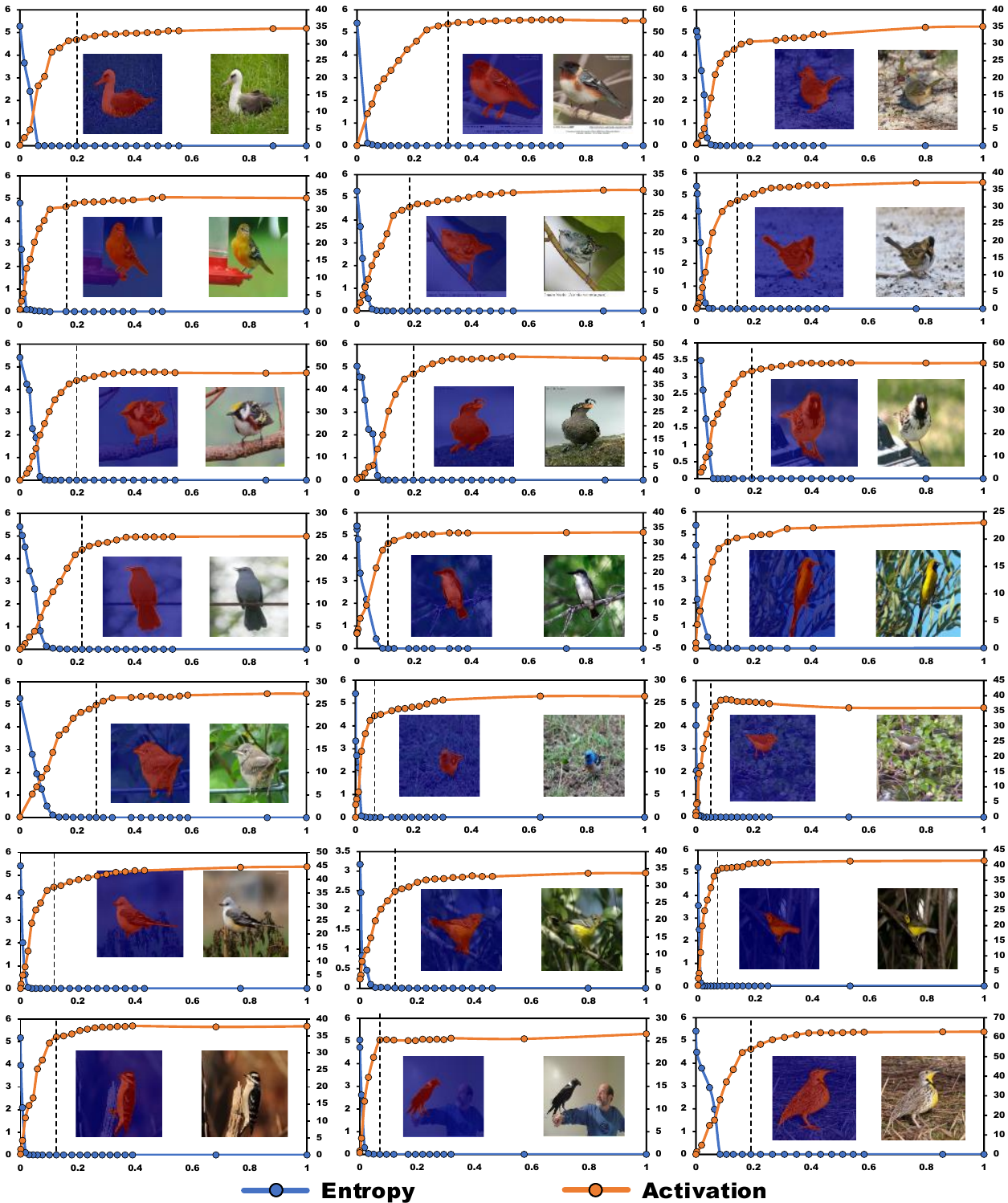}
	\end{overpic}
	\caption{Examples about the entropy value of CE loss $w.r.t$ foreground mask and foreground activation value $w.r.t$ foreground mask.}
	\label{CEvsFG}
\end{figure*}

\begin{figure*}[h]
	\centering
		\begin{overpic}[width=0.95\linewidth]{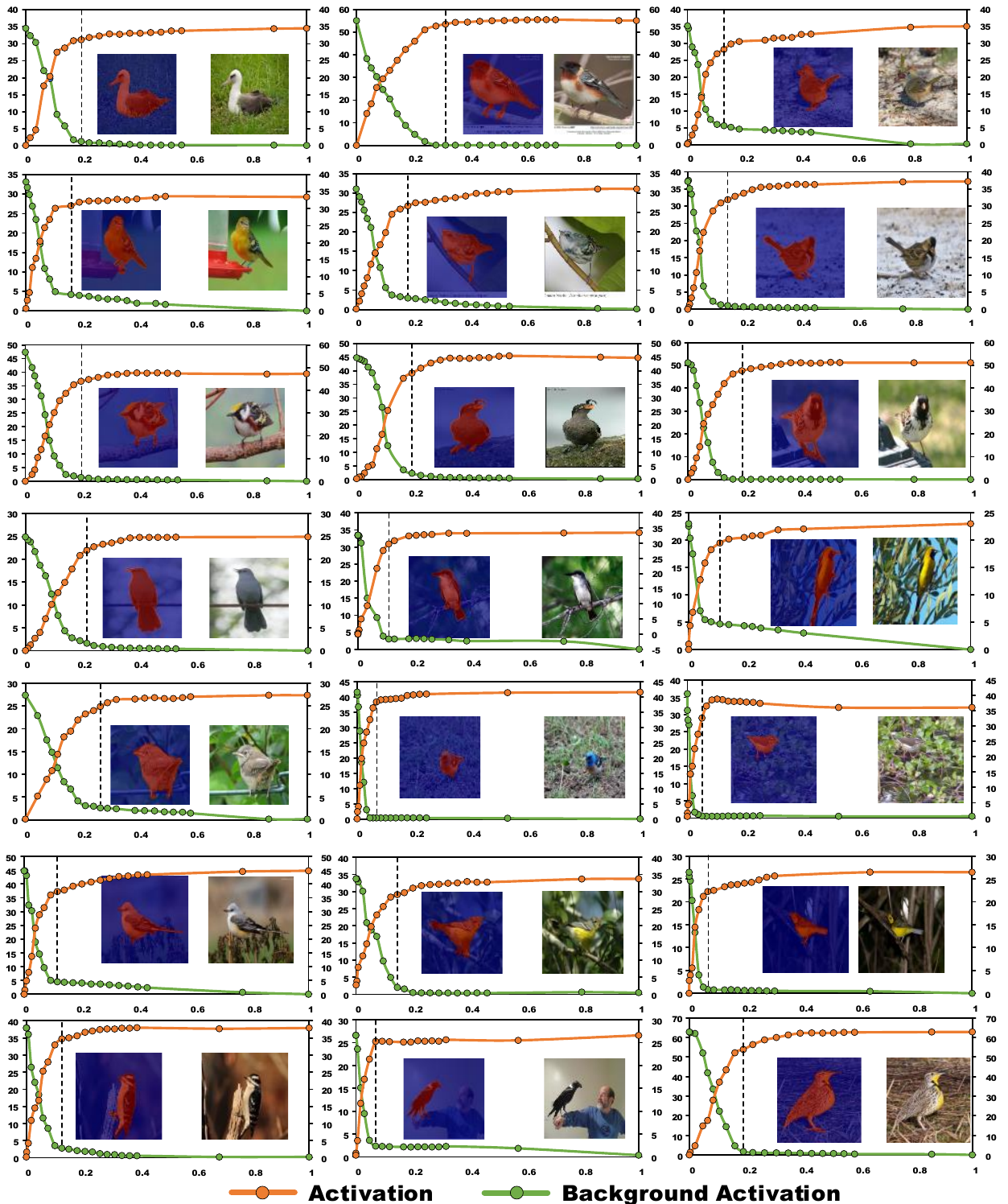}
	\end{overpic}
	\caption{Examples about foreground activation value $w.r.t$ foreground mask and background activation value $w.r.t$ foreground mask.}
	\label{FGvsBG}
\end{figure*}

We introduce the implementation of the experiment, as shown in Fig. \ref{mask_network}(A). For a given GT binary mask, the activation value (Activation) and cross-entropy (Entropy) corresponding to this mask are generated by masking the feature map. We erode and dilate the ground-truth mask with a convolution of kernel size $5n \times 5n$, obtain foreground masks with different area sizes by changing the value of $n$, and plot the activation value versus cross-entropy with the area as the horizontal axis, as shown in Fig. \ref{mask_network}(B). By inverting the foreground mask, the corresponding background activation value for the foreground mask area is generated in the same way. In Fig. \ref{mask_network}(C), we show the curves of entropy, foreground activation, and background activation with mask area. It can be noticed that both background activation and foreground activation value have a higher correlation with the mask compared to the entropy. We show more examples in Fig. \ref{CEvsFG} and Fig. \ref{FGvsBG} to illustrate the generality of this phenomenon. Fig. \ref{CEvsFG} reflects that there is a ``\textbf{mismatch}'' between entropy and ground-truth mask, while the activation value tends to ``\textbf{saturate}'' when mask expands to the object boundary. In Fig. \ref{FGvsBG}, we compare the foreground and background activation curves, which show a ``\textbf{symmetry}", indicating that using the background activation value to learn generator is equally effective.

\section{Experiment}
\subsection{Hyperparameter}
\myPara{Hyperparameter $\alpha$ in total loss.}
$\alpha$ denotes the factor of $\mathcal{L}_{FRG}$. Foreground region guidance loss can guide the activation map learning to the approximate location, which is necessary when the backbone is ResNet50, MobileNetV1, and InceptionV3, but is not required for VGG16. As shown in Table \ref{table:alpha}, on CUB-200-2011, the best results are obtained at $\alpha = 0$ when the backbone is VGG16.

\begin{table*}[h]
\renewcommand{\arraystretch}{1.}
\renewcommand{\tabcolsep}{4.pt}
\centering
\begin{tabular}{c||c|c|c|c|c|c|c|c|c|c|c}
\Xhline{2.\arrayrulewidth}
\hline 
\bm{$\alpha$} &\bm{$0.0$} &\bm{$0.1$} &\bm{$0.2$} &\bm{$0.3$} &\bm{$0.4$} &\bm{$0.5$} &\bm{$0.6$} &\bm{$0.7$}&\bm{$0.8$}&\bm{$0.9$}&\bm{$1.0$} \\
\Xhline{2.\arrayrulewidth}
\hline 
Top-1 &\bm{$71.33$} &$69.79$ &$69.14$ &$69.75$ &$70.07$ &$69.20$ &$70.01$&$69.55$&$69.34$&$69.22$&$68.84$ \\
Top-5 &\bm{$85.33$ }&$83.57$ &$83.00$ &$83.86$ &$83.41$ &$82.74$ &$83.91$&$83.32$&$82.83$&$83.75$ &$82.76$ \\
GT-known &\bm{$91.07$} &$89.24$ &$88.95$ &$89.75$ &$89.33$ &$88.72$ &$89.37$&$89.17$ &$88.67$&$89.70$&$88.64$\\
\hline
\Xhline{2.\arrayrulewidth}
\end{tabular}
\caption{\textbf{Performance $w.r.t$ \bm{$\alpha$}  on CUB-200-2011.}}
\label{table:alpha}
\end{table*}

\myPara{Hyperparameter $\beta$ in total loss.}
$\beta$ reflects the degree of constraint between foreground area and background suppression. when $\beta$ is small, more areas in the foreground activation map are activated, while when $\beta$ is too large, it will suppress the learning of the activation map. As shown in Table \ref{table:beta}, our method achieves the best performance when $\beta$ = 0.7 on VGG16.

\begin{table*}[h]
\renewcommand{\arraystretch}{1.}
\renewcommand{\tabcolsep}{4.pt}
\centering
\begin{tabular}{c||c|c|c|c|c|c|c|c}
\Xhline{2.\arrayrulewidth}
\hline 
\bm{$\beta$} &\bm{$0.1$} &\bm{$0.5$} &\bm{$0.6$} &\bm{$0.7$} &\bm{$0.8$} &\bm{$1.0$} &\bm{$1.2$} &\bm{$1.5$} \\
\Xhline{2.\arrayrulewidth}
\hline 
Top-1 &$68.65$ &$70.35$ &$70.47$ &\bm{$71.33$} &$70.01$ &$70.66$ &$70.13$&$68.90$ \\
Top-5 &$82.32$ &$82.32$ &$84.78$ &\bm{$85.33$} &$84.11$ &$84.73$ &$84.00$&$83.01$ \\
GT-known &$87.79$ &$90.01$ &$90.33$ &\bm{$91.07$} &$90.01$ &$90.30$ &$89.74$&$88.74$ \\
\hline
\Xhline{2.\arrayrulewidth}
\end{tabular}
\caption{\textbf{Performance $w.r.t$ \bm{$\beta$} on CUB-200-2011.}}
\label{table:beta}
\end{table*}

\myPara{Selection of hyperparameter.}
We show the selection of hyperparameters and the corresponding localization accuracy of the proposed BAS for different backbones and datasets in Table \ref{table:hyperparameters}.

\begin{table*}[h]
\renewcommand{\arraystretch}{1.}
\renewcommand{\tabcolsep}{4.pt}
\centering
\begin{tabular}{c|c|cccc||ccc}
\Xhline{2.\arrayrulewidth}
\hline 
Dataset & Backbone & \bm{$\alpha$} & \bm{$\beta$} & \bm{$\lambda$} & K &\textbf{Top-1}&\textbf{Top-5}&\textbf{GT-known}\\
\Xhline{2.\arrayrulewidth}
\hline 
\multirow{4}{*}{CUB-200-2011 \cite{wah2011caltech}} & VGG16 &$0.0$ &$0.7$ & $1.0$ & $80$ & $71.33$&$85.33$ & $91.07$\\
 & MobileNetV1 &$0.5$ & $1.5$ & $1.0$  & $200$ & $69.77$ & $86.00$ & $92.35$ \\
 & ResNet50 &$0.5$ & $1.2$ & $1.0$ & $200$ & $77.25$ & $90.08$ & $95.13$  \\
 & InceptionV3 &$0.5$ & $1.0$ & $1.0$ & $200$ & $73.29$ & $86.31$ & $92.24$  \\
\hline 
\multirow{4}{*}{ILSVRC \cite{russakovsky2015imagenet}} & VGG16 & 0.05 & 1.0 & 1.0 & 1 &52.96 & 65.41 & 69.64\\
 & MobileNetV1 & 0.5  & 1.5 & 1.0 & 1 & 52.97 & 66.59 & 72.00\\
 & ResNet50 & 1.0 & 2.0 & 1.0 & 1 & 57.18 & 68.44 & 71.77\\
 & InceptionV3 & 1.0 & 2.5 & 1.0 & 1 & 58.51 & 67.00 & 71.93 \\
\hline
\Xhline{2.\arrayrulewidth}
\end{tabular}
\caption{\textbf{Selection of hyperparameters under different backbones and datasets and corresponding localization accuracy.}}
\label{table:hyperparameters}
\end{table*}

\myPara{Accuracy.}
We show more results in Table \ref{table:CUB_cls} and Table \ref{table:ILSVRC_cls}. It can be noted that the proposed method achieves excellent localization accuracy along with high accuracy on the classification task, which indicates that BAS can learn object localization results without affecting the classification ability.

\section{More Examples}
\begin{figure*}[t]
	\centering
		\begin{overpic}[width=0.95\linewidth]{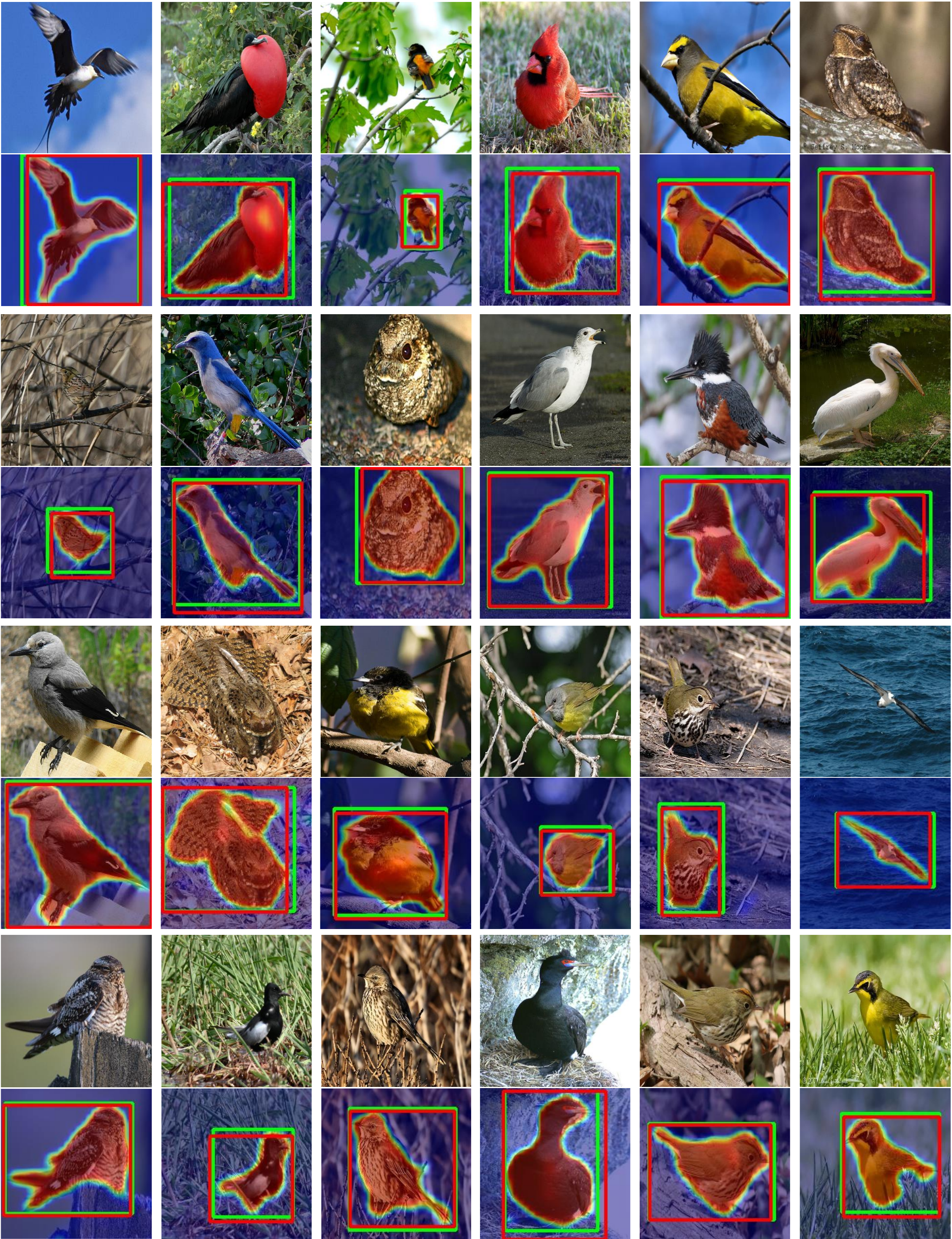}
	\end{overpic}
	\caption{Visualization of the localization results on CUB-200-2011 \cite{wah2011caltech}. The ground-truth bounding boxes are in \textcolor{red}{red}, and the predictions are in \textcolor{green}{green}.}
	\label{CUB}
\end{figure*}

\begin{figure*}[t]
	\centering
		\begin{overpic}[width=0.95\linewidth]{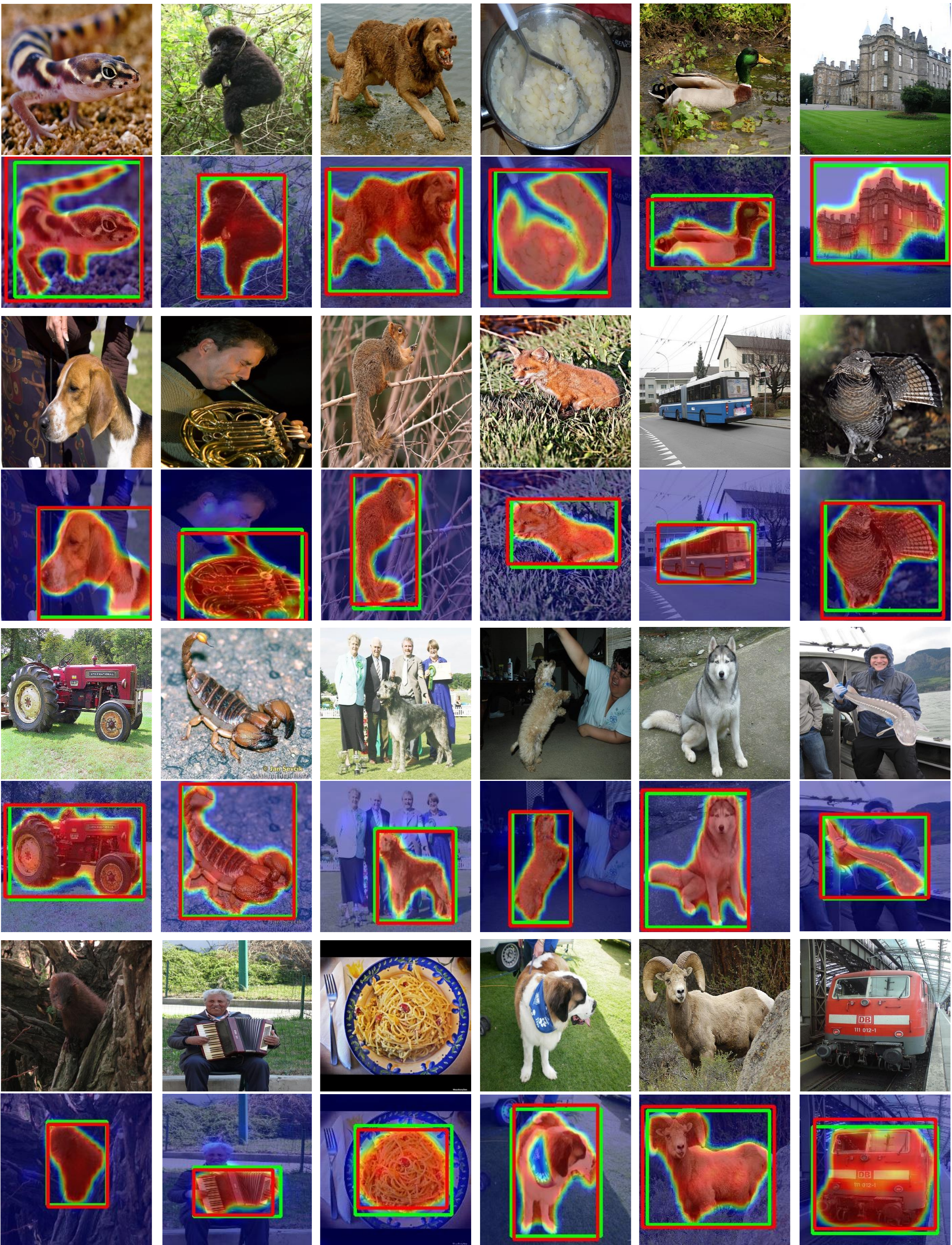}
	\end{overpic}
	\caption{Visualization of the localization results on ILSVRC \cite{russakovsky2015imagenet}. The ground-truth bounding boxes are in \textcolor{red}{red}, and the predictions are in \textcolor{green}{green}.}
	\label{ILSVRC}
\end{figure*}

\myPara{Visual Results.}
More visualization examples are shown in Fig. \ref{CUB} and Fig. \ref{ILSVRC}. As can be seen in Fig. \ref{ILSVRC}, even in a noisy environment, BAS can still accurately localize objects, which indicates that the proposed BAS has robust localization capability.

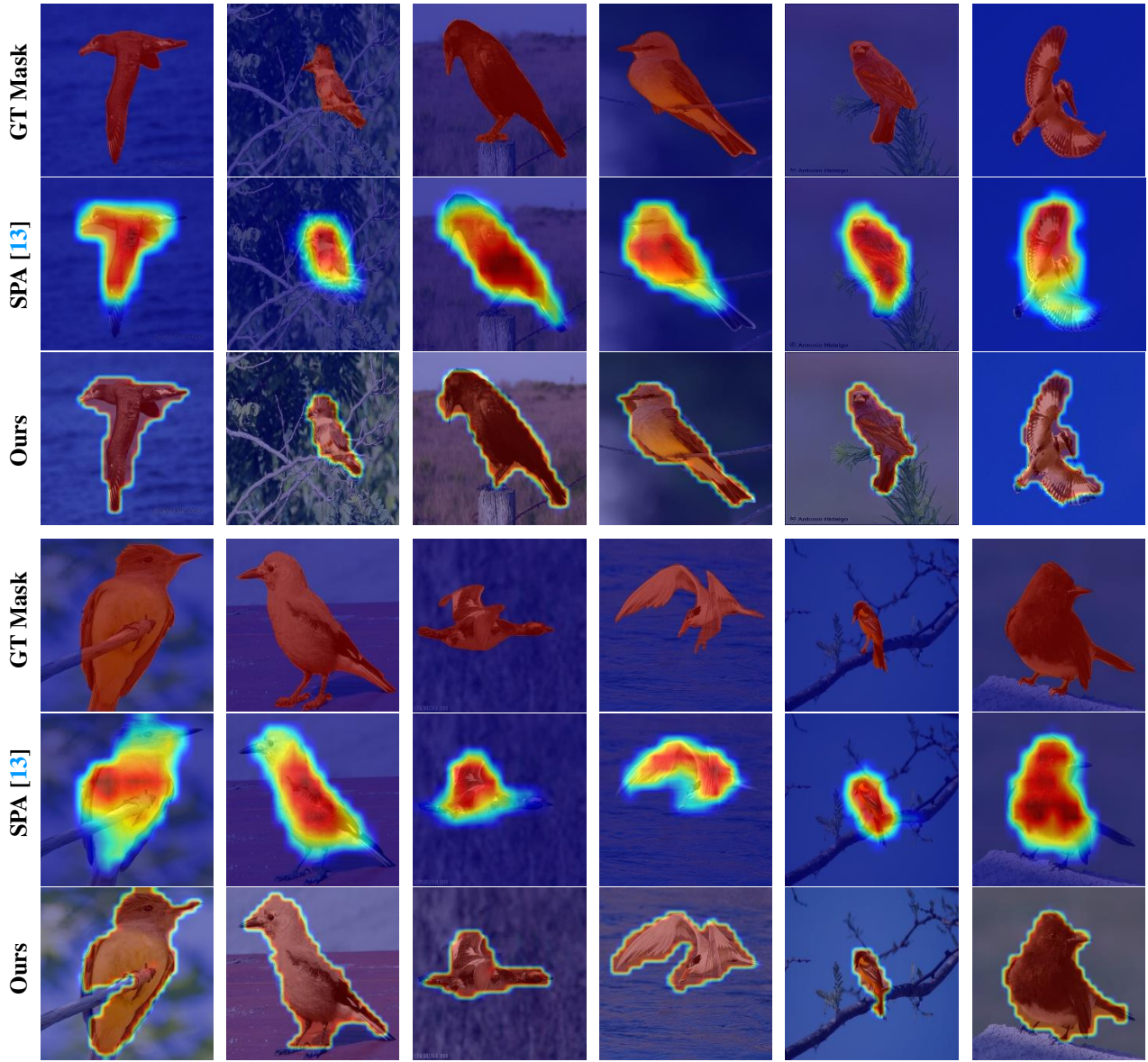
\begin{figure*}[t]
	\centering
		\begin{overpic}[width=0.95\linewidth]{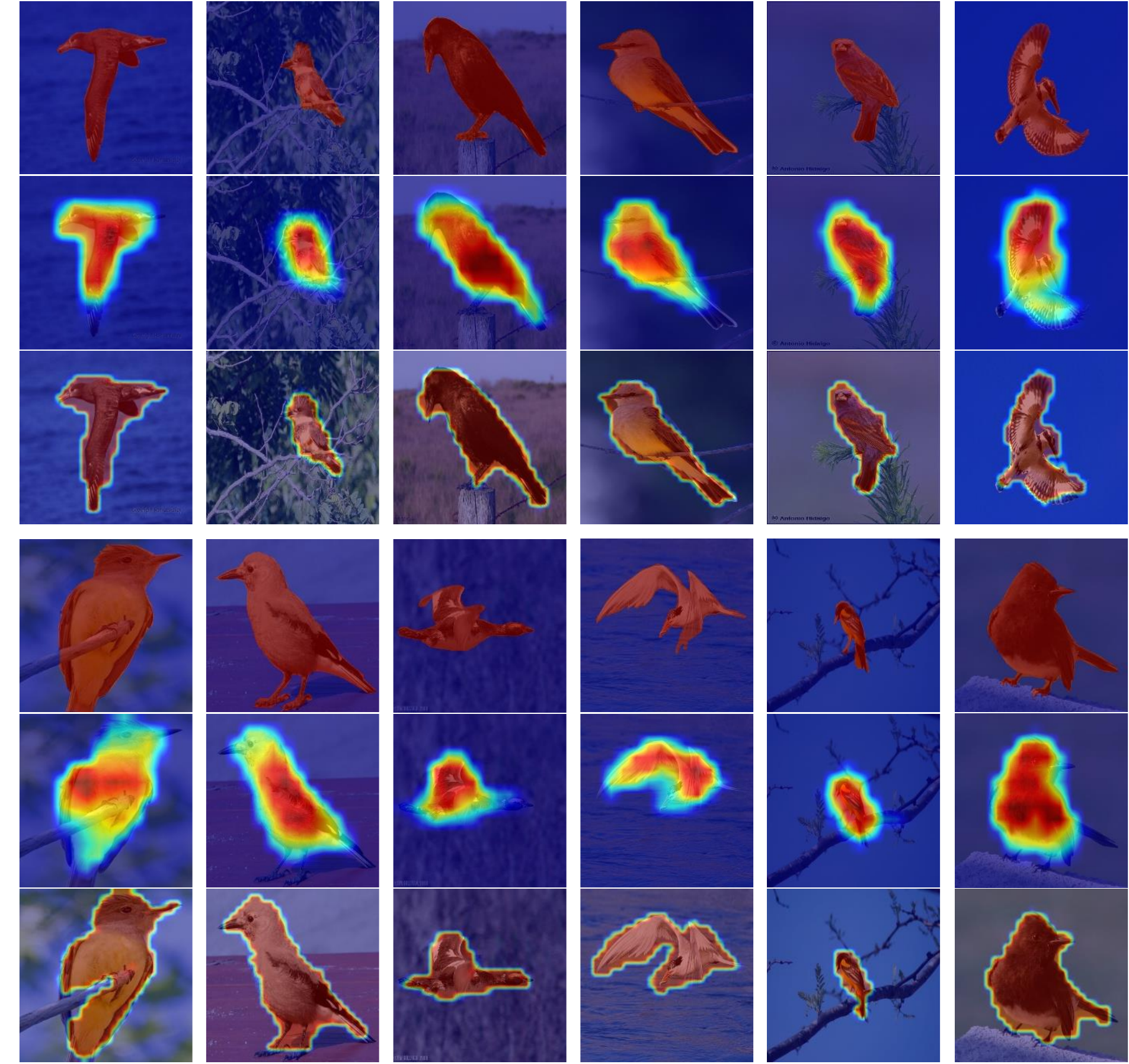}
		\put(-1,35){\rotatebox{90}{\textbf{GT Mask}}}
	    \put(-1,20){\rotatebox{90}{\textbf{SPA \cite{pan2021unveiling}}}}
	    \put(-1,6){\rotatebox{90}{\textbf{Ours}}}
	    
	    \put(-1,82){\rotatebox{90}{\textbf{GT Mask}}}
	    \put(-1,67){\rotatebox{90}{\textbf{SPA \cite{pan2021unveiling}}}}
	    \put(-1,53){\rotatebox{90}{\textbf{Ours}}}
	\end{overpic}
	\caption{Illustration of the mask provided by CUB-200-2011, and comparison of the localization maps of SPA and our method on VGG16.}
	\label{Mask}
\end{figure*}

\myPara{Mask Annotation.}
We demonstrate part of the mask labels provided by CUB-200-2011 and compare the localization maps of SPA \cite{pan2021unveiling} and BAS on VGG16, as shown in Fig. \ref{Mask}. Compared to SPA, our localization maps are brighter on the object area and better localized at the edges of the object.

\begin{table*}[t]
\renewcommand{\arraystretch}{1.}
\renewcommand{\tabcolsep}{8pt}
\centering
\begin{tabular}{l||c||c||c|c|c||c|c}
\Xhline{2.0\arrayrulewidth}
\hline
\multirow{2}{*}{\textbf{Methods}} & \multirow{2}{*}{\textbf{Venue}} & \multirow{2}{*}{\textbf{Backbone}} &  \multicolumn{3}{c||}{\textbf{Loc. Acc.}} & \multicolumn{2}{c}{\textbf{Cls. Acc.}} \\
\cline{4-8}
      & &  & \textbf{Top-1} & \textbf{Top-5} & \textbf{GT-known}  & \textbf{Top-1} & \textbf{Top-5}  \\
\hline
\Xhline{2.0\arrayrulewidth}
CAM~\cite{zhou2016learning} & CVPR$16$ &VGG16
&$41.06$&$50.66$& $55.10$  & $76.60$&$92.50$  \\
ACoL~\cite{zhang2018adversarial} & CVPR$18$&VGG16
&$45.92$&$56.51$& $62.96$ &$71.90$&$-$\\
ADL~\cite{choe2019attention} & CVPR$19$&VGG16
&$52.36$&- & $75.41$ &$65.27$&$-$\\
DANet~\cite{xue2019danet} & ICCV$19$ &VGG16
&$52.52$&$61.96$&$67.70$ &$75.40$&$92.30$\\
I2C~\cite{zhang2020inter} & ECCV$20$ &VGG16
&$55.99$&$68.34$&$-$ & $-$ & $-$\\
MEIL~\cite{mai2020erasing} & CVPR20 &VGG16
&$57.46$&$-$&$73.84$ & $74.77$ &$-$\\
GCNet~\cite{lu2020geometry} & ECCV$20$ &VGG16
&$63.24$&$75.54$&$81.10$ &$76.80$&$92.30$\\
PSOL~\cite{zhang2020rethinking} & CVPR$20$ &VGG16
& $66.30$&\underline{$84.05$}&$89.11$ & $-$&$-$\\
SPA~\cite{pan2021unveiling} & CVPR$21$ &VGG16
&$60.27$ &$72.50$&$77.29$ & $76.11$&$92.15$\\
SLT~\cite{guo2021strengthen} & CVPR$21$ &VGG16
& $67.80$&$-$&$87.60$ & $76.60$&$-$\\
FAM~\cite{meng2021foreground} & ICCV$21$ &VGG16
&\underline{$69.26$} &$-$&\underline{$89.26$} & \underline{$77.26$}&$-$\\
ORNet~\cite{xie2021online} & ICCV$21$ &VGG16
&$67.73$&$80.77$&$86.20$ & $77.00$&\underline{$93.00$}\\
\hline
\rowcolor{mygray}
\textbf{\texttt{BAS(Ours)}}&This Work &VGG16&
\bm{$71.33 $}& \bm{$85.33 $}&\bm{$91.07 $} & \bm{$77.49$}& \bm{$93.18$}\\
\hline
\Xhline{1.5\arrayrulewidth}
CAM~\cite{zhou2016learning} & CVPR$16$  &MobileNetV1
&$48.07$&\underline{$59.20$}&$63.30$&$73.25$&\underline{$91.50$}\\
HaS~\cite{singh2017hide} & ICCV$17$ &MobileNetV1
&$46.70$&$-$&$67.31$ &$65.98$&$-$\\
ADL~\cite{choe2019attention} & CVPR$19$ &MobileNetV1
&$47.74$&$-$&$-$ &$70.43$&$-$\\
RCAM~\cite{bae2020rethinking} & ECCV$20$ &MobileNetV1
&$59.41$&$-$&$78.60$ &$73.51$&$-$\\
FAM~\cite{meng2021foreground} & ICCV$21$ &MobileNetV1
&\underline{$65.67$}&$-$&\underline{$85.71$} &\bm{$76.38$}&$-$\\
\hline
\rowcolor{mygray}
\textbf{\texttt{BAS(Ours)}} & This Work & MobileNetV1
& \bm{$69.77 $}& \bm{$86.00 $}&\bm{$92.35 $} & \underline{$74.67$} &\bm{$92.60$}\\
\hline
\Xhline{1.5\arrayrulewidth}
CAM~\cite{zhou2016learning} & CVPR$16$ &ResNet50
&$46.71$&$54.44$&$57.35$ & $80.26$ & $-$\\
ADL~\cite{choe2019attention} & CVPR$19$ &ResNet50-SE
&$62.29$&$-$&$-$ & $80.34$ & $-$ \\
PSOL~\cite{zhang2020rethinking} & CVPR$20$ &ResNet50
& $70.68$&$86.64$&$90.00$ &  $-$ & $-$ \\
WTL~\cite{babar2021look} & WACV$21$ &ResNet50
&$64.70$&$-$&$77.35$ & $77.28$ & $-$ \\
FAM~\cite{meng2021foreground} & ICCV$21$ &ResNet50
&$73.74$&$-$&$85.73$ &\bm{ $82.72$} & $-$ \\
SPOL~\cite{wei2021shallow} & CVPR$21$&ResNet50
& \bm{$80.12$}&\bm{$93.44$}&\bm{$96.46$} & $-$ & $-$ \\
\hline
\rowcolor{mygray}
$\textbf{\texttt{BAS(Ours)}}$ & This Work &ResNet50
&\underline{$77.25 $}&\underline{$90.08 $}&\underline{$95.13 $} & \underline{$80.84$ }& \bm{$94.39$} \\
\hline
\Xhline{1.5\arrayrulewidth}
CAM~\cite{zhou2016learning} & CVPR$16$ &InceptionV3
&$41.06$&$50.66$&$55.10$ &  $73.80$ & $91.50$ \\
SPG~\cite{zhang2018self} & ECCV$18$ &InceptionV3
&$46.64$&$57.72$&$-$ & $-$ & $-$ \\
DANet~\cite{xue2019danet} & ICCV$19$ &InceptionV3
&$49.45$&$60.46$&$67.03$ & $71.20$ & $90.60$ \\
I2C~\cite{zhang2020inter} & ECCV$20$ &InceptionV3
&$55.99$&$68.34$&$72.60$ & $-$ & $-$ \\
GCNet~\cite{lu2020geometry} & ECCV$20$ &InceptionV3
&$58.58$&$71.00$&$75.30$ & $76.80$ & \bm{$93.40$} \\
PSOL~\cite{zhang2020rethinking} & CVPR$20$ &InceptionV3
& $65.51$&\underline{$83.44$}&$-$ &  $-$ & $-$ \\
SPA~\cite{pan2021unveiling} & CVPR$21$ &InceptionV3
&$53.59$&$66.50$&$72.14$ & $73.51$ & $91.39$ \\
SLT~\cite{guo2021strengthen} & CVPR$21$&InceptionV3
& $66.10$&$-$&$86.50$ &  $76.40$ & $-$ \\
FAM~\cite{meng2021foreground} & ICCV$21$ &InceptionV3
&\underline{$70.67$}&$-$&\underline{$87.25$} & \bm{$81.25$} & $-$ \\
\hline
\rowcolor{mygray}
$\textbf{\texttt{BAS(Ours)}}$ & This Work &InceptionV3
&\bm{$73.29 $}&\bm{$86.31 $}&\bm{$92.24 $} & \underline{$79.01$} & \underline{$93.10$} \\
\hline
\Xhline{2.0\arrayrulewidth}
\end{tabular}
\caption{Comparison of localization accuracy and classification accuracy with state-of-the-art methods on CUB-200-2011. Best results are highlighted in \textbf{bold}, second are \underline{underlined}.}
\label{table:CUB_cls}
\end{table*}

\begin{table*}[t]
\renewcommand{\arraystretch}{1.}
\renewcommand{\tabcolsep}{8pt}
\centering
\begin{tabular}{l||c||c||c|c|c||c|c}
\Xhline{2.0\arrayrulewidth}
\hline
\multirow{2}{*}{\textbf{Methods}} & \multirow{2}{*}{\textbf{Venue}} & \multirow{2}{*}{\textbf{Backbone}} &  \multicolumn{3}{c||}{\textbf{Loc. Acc.}} & \multicolumn{2}{c}{\textbf{Cls. Acc.}} \\
\cline{4-8}
      & &  & \textbf{Top-1} & \textbf{Top-5} & \textbf{GT-known}  & \textbf{Top-1} & \textbf{Top-5}  \\
\hline
\Xhline{2.0\arrayrulewidth}
CAM~\cite{zhou2016learning} & CVPR$16$ &VGG16
&$42.80$&$54.86$&$59.00$  & $ 66.60$ & $ 88.60$  \\
ACoL~\cite{zhang2018adversarial} & CVPR$18$&VGG16
&$45.83$&$59.43$&$62.96$ & $ 67.50$ & $ 88.00$ \\
ADL~\cite{choe2019attention} & CVPR$19$&VGG16
&$44.92$&$-$&$-$ & $69.48 $ & $- $ \\
I2C~\cite{zhang2020inter} & ECCV$20$ &VGG16
& $47.41$ & $58.51$&$63.90$& $ 69.40$ & $ 89.30$ \\
MEIL~\cite{mai2020erasing} & CVPR20 &VGG16
& $46.81$ &$-$&$-$ & $70.27 $ & $ -$ \\
PSOL~\cite{zhang2020rethinking} & CVPR$20$ &VGG16
& $50.89$&$60.90$&$64.03$ & $- $ & $- $ \\
SPA~\cite{pan2021unveiling} & CVPR$21$ &VGG16
& $49.56$&$61.32$&$65.05$ & $70.51$ & $90.05$ \\
SLT~\cite{guo2021strengthen} & CVPR$21$ &VGG16
& $51.20$&$62.40$&$67.20$ & \bm{$72.40$} & $- $ \\
FAM~\cite{meng2021foreground} & ICCV$21$ &VGG16
& $51.96$&$-$&\bm{$71.73$} & $70.90$ & $ -$ \\
ORNet~\cite{xie2021online} & ICCV$21$ &VGG16
& \underline{$52.05$}&\underline{$63.94$}&$68.27$ & \underline{$71.60$} & \underline{$90.40$} \\
\hline
\rowcolor{mygray}
\textbf{\texttt{BAS(Ours)}}&This Work &VGG16
& \bm{$52.96 $}& \bm{$65.41 $}&\underline{$69.64 $}& $70.84$ & \bm{$90.46$} \\
\hline
\Xhline{1.5\arrayrulewidth}
CAM~\cite{zhou2016learning} & CVPR$16$  &MobileNetV1
&$43.35$&$\underline{54.44}$&$58.97$ & $66.20$ & \underline{$87.23$}\\
HaS~\cite{singh2017hide} & ICCV$17$ &MobileNetV1
&$42.73$&$-$&$60.12$ & $65.45$ & $- $\\
ADL~\cite{choe2019attention} & CVPR$19$ &MobileNetV1
&$43.01$&$-$&$-$ & $67.77$ & $ -$\\
RCAM~\cite{bae2020rethinking} & ECCV$20$ &MobileNetV1
&$44.78$&$-$&$61.69$ & $67.15$ & $- $\\
FAM~\cite{meng2021foreground} & ICCV$21$ &MobileNetV1
&\underline{$46.24$}&$-$&\underline{$62.05$} & \bm{$70.28$} & $-$\\
\hline
\rowcolor{mygray}
\textbf{\texttt{BAS(Ours)}} & This Work & MobileNetV1
& \bm{$52.97 $}& \bm{$66.59 $}&\bm{$72.00 $} & \underline{$68.94$} & \bm{$89.28$}\\
\hline
\Xhline{1.5\arrayrulewidth}
CAM~\cite{zhou2016learning} & CVPR$16$ &ResNet50
 & $38.99$&$49.47$&$51.86$ & $- $ & $ -$\\
ADL~\cite{choe2019attention} & CVPR$19$ &ResNet50-SE
&$48.53$&$-$&$-$ & $75.85$ & $-$\\
I2C~\cite{zhang2020inter} & ECCV$20$ &ResNet50
& $51.83$&$64.60$&$68.50$ & \bm{$76.70$} & $- $\\
PSOL~\cite{zhang2020rethinking} & CVPR$20$ &ResNet50
& $53.98$&$63.08$&$65.44$ & $- $ & $ -$\\
WTL~\cite{babar2021look} & WACV$21$ &ResNet50
&$52.36$&$-$&$67.89$ & $-$ & $-$ \\
FAM~\cite{meng2021foreground} & ICCV$21$ &ResNet50
 &$54.46$&$-$&$64.56$ & \underline{$76.48$} & $-$\\
SPOL~\cite{wei2021shallow} & CVPR$21$&ResNet50
&\bm{$59.14$}&\underline{$67.15$}&\underline{$69.02$} & $-$ & $-$\\
\hline
\rowcolor{mygray}
$\textbf{\texttt{BAS(Ours)}}$ & This Work &ResNet50
&\underline{$57.18 $}&\bm{$68.44 $}&\bm{$71.77 $} & $76.06$ & \bm{$93.12$}\\
\hline
\Xhline{1.5\arrayrulewidth}
CAM~\cite{zhou2016learning} & CVPR$16$ &InceptionV3
 & $46.29$&$58.19$&$62.68$  & $73.30$ & $91.80$\\
SPG~\cite{zhang2018self} & ECCV$18$ &InceptionV3
&$48.60$&$60.00$&$64.69$ & $69.70$ & $90.10 $\\
DANet~\cite{xue2019danet} & ICCV$19$ &InceptionV3
&$47.53$&$58.28$&$-$ & $72.50$ & $91.40$\\
I2C~\cite{zhang2020inter} & ECCV$20$ &InceptionV3
&$53.11$&$64.13$&$68.50$ & $73.30$ & $91.60$\\
GCNet~\cite{lu2020geometry} & ECCV$20$ &InceptionV3
&$49.06$&$58.09$&$-$ & $77.40$ & \underline{$93.60$}\\
PSOL~\cite{zhang2020rethinking} & CVPR$20$ &InceptionV3
 & $54.82$&$63.25$&$65.21$ & $-$ & $-$\\
SPA~\cite{pan2021unveiling} & CVPR$21$ &InceptionV3
&$52.73$&$64.27$&$68.33$ & $73.26$ & $91.81 $\\
SLT~\cite{guo2021strengthen} & CVPR$21$&InceptionV3
& \underline{$55.70$}&\underline{$65.40$}&$67.60$ & \bm{$78.10$} & $-$\\
FAM~\cite{meng2021foreground} & ICCV$21$ &InceptionV3
&$55.24$&$-$&\underline{$68.62$} & $77.63$ & $-$\\
\hline
\rowcolor{mygray}
$\textbf{\texttt{BAS(Ours)}}$ & This Work &InceptionV3
&\bm{$58.51 $}&\bm{$69.00 $}&\bm{$71.93 $} & \underline{$77.99$} & \bm{$94.02$}\\
\hline
\Xhline{2.0\arrayrulewidth}
\end{tabular}
\caption{Comparison of localization accuracy and classification accuracy with state-of-the-art methods on ILSVRC. Best results are highlighted in \textbf{bold}, second are \underline{underlined}.}
\label{table:ILSVRC_cls}
\end{table*}

\clearpage

\end{document}